\documentclass[lettersize,journal]{IEEEtran}
\usepackage{amsmath,amsfonts}
\usepackage{algorithmic}
\usepackage{algorithm}
\usepackage{array}
\usepackage{textcomp}
\usepackage{stfloats}
\usepackage{url}
\usepackage{verbatim}
\usepackage{graphicx}
\usepackage{cite}
\usepackage{caption}
\usepackage{subcaption}
\usepackage{bbding}

\usepackage{tabularx}
\usepackage[table]{xcolor}
\usepackage{colortbl}
\usepackage{tablestyles}
\usepackage{multicol}
\usepackage{multirow}
\usepackage{bigstrut}

\usepackage[utf8]{inputenc}
\usepackage{pgfplots}
\DeclareUnicodeCharacter{2212}{−}
\usepgfplotslibrary{groupplots,dateplot}
\usetikzlibrary{patterns,shapes.arrows}
\pgfplotsset{compat=newest}
\usepackage{pdfpages}

\usepackage[font=normalfont,labelfont=bf]{caption}
\usepackage{balance}
\begin{document}

%\title{Latent in the Wild: Database, Evaluation Platform, and Benchmarking}
%\title{Benchmarking Latent in the Wild with a new 13K Latent Fingerprint Database}
\title{A Latent Fingerprint in the Wild Database}

\author{Xinwei Liu, Kiran Raja, Renfang Wang, Hong Qiu, Hucheng Wu, Dechao Sun, Qiguang Zheng, Nian Liu, Xiaoxia Wang, Gehang Huang, Raghavendra Ramachandra, Christoph Busch 
%\author{IEEE Publication Technology,~\IEEEmembership{Staff,~IEEE,}
        % <-this % stops a space
\thanks{This paper was supported by the National Natural Science Foundation of China (Grant No. 62106228, 61906170), Zhejiang Natural Science Foundation (Grant No. LQ22F020003), Ningbo Natural Science Foundation (Grant No. 2021J175), and the Ningbo Yongjiang Talent Introduction Programme 2021.}% <-this % stops a space
\thanks{Xinwei Liu, Renfang Wang, Hong Qiu, Hucheng Wu, Qiguang Zheng are with the College of Big Data and Software Engineering, Zhejiang Wanli University (ZWU), 8 Qianhu South Road, Ningbo, Zhejiang, China.}% <-this % stops a space
\thanks{Kiran Raja, Raghavendra Ramachandra, Christoph Busch are with  Norwegian University of Science and Technology (NTNU), 2815 Gj\o vik, Norway.}% <-this % stops a space
\thanks{Dechao Sun is with the College of Digital Technology and Engineering, Ningbo University of Finance \& Economics, Ningbo, China.}% <-this % stops a space
% \thanks{Matteo Ferrara is with the Department of Computer Science and Engineering, University of Bologna, 47521 Cesena, Italy.}% <-this % stops a space
% \thanks{Philipp Terh{\"{o}}rst and Andr{\'{e}} Boller are with the Fraunhofer Institute for Computer Graphics Research IGD, Darmstadt, Germany.}% <-this % stops a space
\thanks{Nian Liu and Xiaoxia Wang are with the Library and Learning Resource Center, Zhejiang Wanli University (ZWU), 8 Qianhu South Road, Ningbo, Zhejiang, China.}% <-this % stops a space
\thanks{Gehang Huang is with the HDU-ITMO Joint Institute, Hangzhou Dianzi University, Hangzhou, China.}% <-this % stops a space
\thanks{Manuscript received XXXX XX, 2023; revised XXXX XX, 2023.}
\renewcommand\thefigure{\arabic{figure}}
\setcounter{figure}{0}
\begin{center}
\centering
\includegraphics[width=0.8\linewidth]{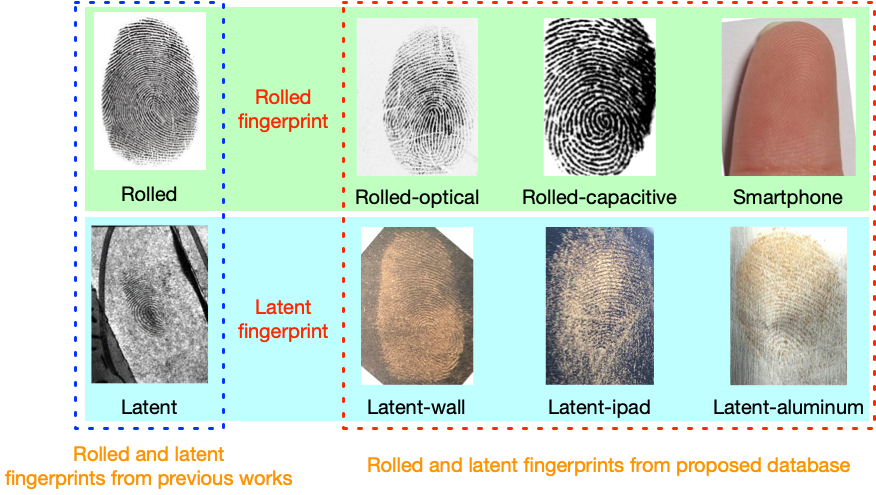}
\captionof{figure}{Our contribution of Latent Fingerprint In the Wild (LFIW) dataset compared to previous works.}
\label{fig:fingerprint_examples}
\end{center}
}

\markboth{Journal of \LaTeX\ Class Files,~Vol.~XX, No.~XX, XXXX~2023}%
{Shell \MakeLowercase{\textit{et al.}}: A Sample Article Using IEEEtran.cls for IEEE Journals}


\maketitle
\begin{abstract}
Latent fingerprints are among the most important and widely used evidence in crime scenes, digital forensics and law enforcement worldwide. Despite the number of advancements reported in recent works, we note that significant open issues such as independent benchmarking and lack of large-scale evaluation databases for improving the algorithms are inadequately addressed. The available databases are mostly of semi-public nature, lack of acquisition in the wild environment, and post-processing pipelines. Moreover, they do not represent a realistic capture scenario similar to real crime scenes, to benchmark the robustness of the algorithms. Further, existing databases for latent fingerprint recognition do not have a large number of unique subjects/fingerprint instances or do not provide ground truth/reference fingerprint images to conduct a cross-comparison against the latent. In this paper, we introduce a new wild large-scale latent fingerprint database that includes five different acquisition scenarios: reference fingerprints from (1) optical and (2) capacitive sensors, (3) smartphone fingerprints, latent fingerprints captured from (4) wall surface, (5) Ipad surface, and (6) aluminium foil surface. The new database consists of 1,318 unique fingerprint instances captured in all above mentioned settings. A total of 2,636 reference fingerprints from optical and capacitive sensors, 1,318 fingerphotos from smartphones, and 9,224 latent fingerprints from each of the 132 subjects were provided in this work. The dataset is constructed considering various age groups, equal representations of genders and backgrounds. In addition, we provide an extensive set of analysis of various subset evaluations to highlight open challenges for future directions in latent fingerprint recognition research.
\end{abstract}

% In addition, a new and independent online evaluation platform is also presented to facilitate reproducible research where any researcher, governmental agency or private entity can submit and evaluate their latent fingerprint recognition methods.


\section{Introduction}
\IEEEPARstart{L}{atent} fingerprints were first reported to convict a suspect as evidence in 1893 \cite{cao2018automated}. Over the years, latent fingerprints have been regarded as one of the most commonly and broadly used sources of evidence in crime scenes, digital forensics, law enforcement, etc \cite{cao2018automated}. Latent fingerprints can be left on various surfaces when a finger makes contact with an object. The manner in which a finger touches the surface of an object has a significant impact on the latent fingerprint quality (e.g., sharpness, contrast, and visible area). There is a long history that latent fingerprint recognition was performed by latent examiners before the development of Automated Fingerprint Identification System (AFIS). In recent years, latent AFIS has become one of the most commonly used technologies by law enforcement agencies worldwide \cite{maltoni2022latent}. More than 300,000 latent fingerprint identification demands were sent to the FBI over the United States only in 2020 \cite{fbi2020}. 

Unlike rolled and slap fingerprints (reference fingerprint images acquired using standard fingerprint capture devices), latent fingerprints are captured under unconstrained and unsupervised conditions. Low quality, partial visibility, and the absence of satisfactory number and quality of minutiae points are common issues faced in latent fingerprint recognition. The National Institute of Standards  and Technology (NIST) announced two fingerprint vendor technology evaluations (FpVTE) in 2003 and 2012, respectively, \cite{fpvte} to advance the research on latent fingerprint recognition. FpVTE was intended for the evaluation of fingerprint system performance to meet the requirements for real-world applications for both reference and latent fingerprints. In the latest FpVTE2012, the lowest False Negative Identification Rate (FNIR) and False Positive Identification Rate (FPIR) were reported as $1.9\%$ and $0.1\%$ by the top-performing AFIS for reference fingerprints \cite{watson2012fingerprint}. However, the best identification rate was only $67.2\%$ for latent fingerprints during the NIST Evaluation of Latent Fingerprint Technologies: Extended Feature Sets (ELFT-EFS) \cite{indovina2012evaluation}. The major difference in recognition performance between the reference and latent fingerprints is mainly caused by the low fingerprint quality of the ridge-and-valley structures in latent fingerprints. It is obvious that the further development of robust and high-accuracy latent fingerprint recognition systems is necessary, however a progress is currently limited by the sparse access to openly available datasets.

% \begin{figure*}[t]
%     \centering
%     \includegraphics[width=0.6\textwidth]{fingerprint_examples.png}
%     \caption{Examples of rolled and latent fingerprints from previous works and the LFIW database proposed in this paper.}
%     \label{fig:fingerprint_examples}
% \end{figure*}

In the past decade, several studies have focused on developing latent fingerprint recognition algorithms \cite{singla2020automated}. However, the performance evaluation of these methods was conducted on only a few databases, such as the NIST SD27 Database \cite{garris2000nist}, IIIT-D Latent Fingerprint Database \cite{sankaran2011matching}, and Tsinghua Overlapped Latent Fingerprint Database \cite{feng2012robust}. Despite being valuable, these databases have major shortcomings:1) a small number of subjects respectively finger instances and latent fingerprint samples, 2) a constrained acquisition environment, and 3) limited availability. Moreover, one of the most commonly used latent fingerprint databases NIST SD27 Database has been withdrawn, making the development and performance evaluation of latent fingerprint recognition even more difficult.

In this paper, we first review all existing studies on latent fingerprint recognition to provide an overview to the reader. We then provide an extensive analysis of algorithms that are relevant for segmentation, minutiae extraction, and the comparison of latent fingerprints both within and across sensors. From a review of existing works, we note that latent fingerprint recognition algorithms have rarely been tested on large-scale datasets \cite{ICPSR2009}. To the best of our knowledge, there is no large-scale latent fingerprint in the wild database that containing both reference fingerprints (ground truth) and latent fingerprints acquired from different surfaces. Therefore, it is necessary to establish a new large-scale latent fingerprint in the wild database to meet the need for robust latent fingerprint recognition algorithm development and evaluation.

\begin{figure}
    \centering
    \includegraphics[width=\columnwidth]{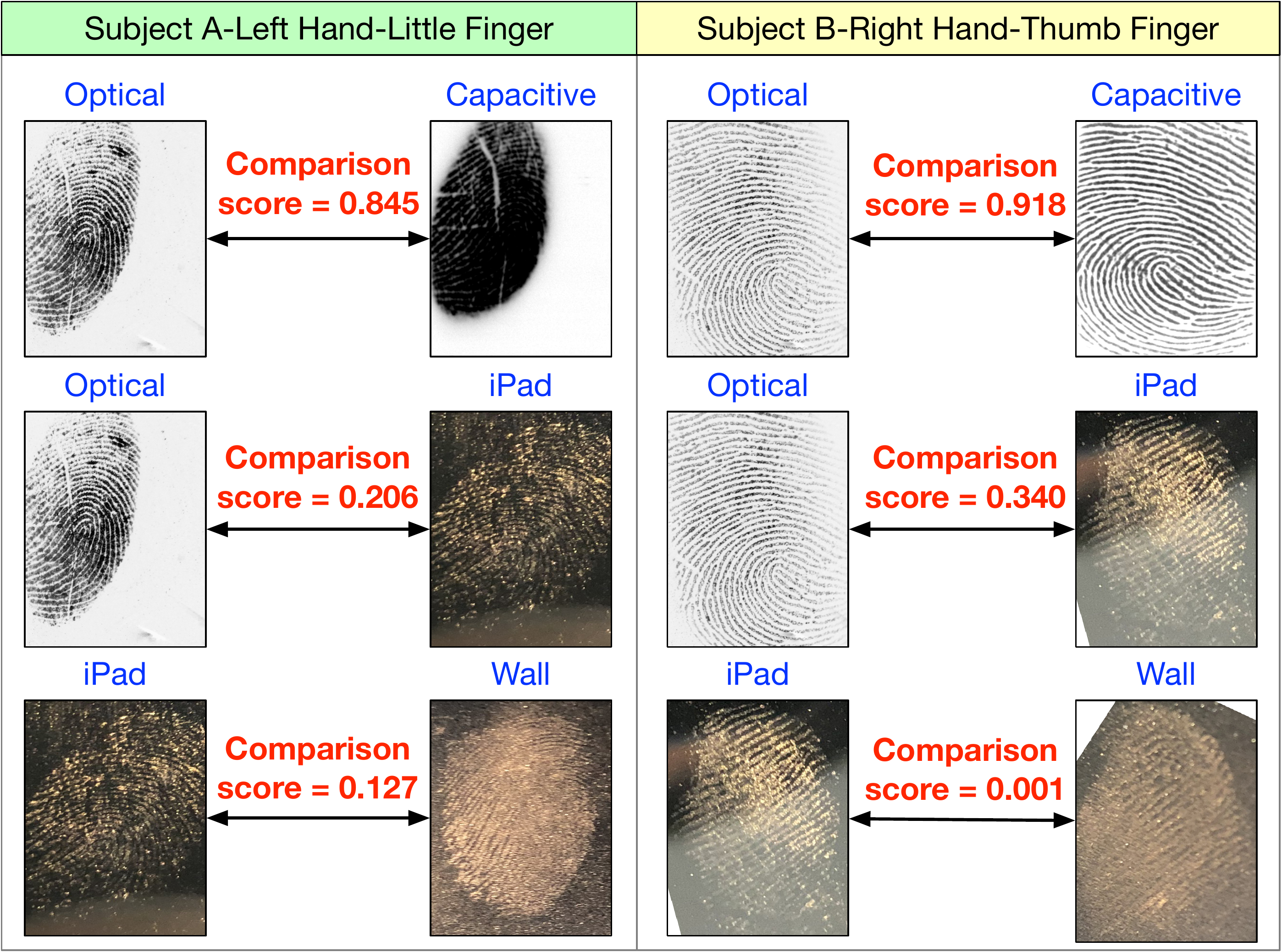}
    \caption{Examples of comparison scores of the fingerprints from the proposed LFIW database.}
    \label{fig:score_drop}
\end{figure}

In order to address the limitations mentioned above, we provide the following three major contributions in this paper:
\begin{itemize}
    \item Noting the non-availability of public datasets, a large-scale database of latent fingerprints in the wild is presented in this work which is referred to as "Latent Fingerprint In the Wild" (LFIW). The dataset is collected in six different scenarios, constituting a total of 13180 images of 132 subjects, and is released along with this paper. This dataset contains various age groups and equal representations of genders and backgrounds, making it a unique dataset for the performance evaluation of latent fingerprint recognition algorithms. As can be seen in Fig. \ref{fig:score_drop}, the comparison scores significantly decrease from reference \textbf{\textit{vs.}} reference comparison to latent \textbf{\textit{vs.}} latent comparison, which also indicates that the LFIW dataset is suitable for the evaluation and development for the current and future latent fingerprint recognition techniques. The LFIW dataset is available for academic research purposes \footnote{Upon acceptance of the paper.}. 
    \item Unlike other works, we also present a ground truth of fingerprints captured using optical and capacitive sensors to conduct an analysis of contact-based versus latent fingerprint comparison. In addition, owing to recent trends in the use of smartphone-based fingerphotos in biometrics, we also introduce fingerphoto images to benchmark the latent to the fingerphoto recognition. 
    \item A benchmark and independent evaluation of 5 state-of-the-art fingerprint recognition methods and 1 latent fingerprint recognition approach is presented to highlight the performance limitations of existing approaches. For each method, a total of 118,620 mated comparison scores and 173,593,780 non-mated comparison scores were generated for performance evaluation to derive statistically significant conclusions

    % \item As a minor contribution, we also present a new and independent evaluation platform to facilitate reproducible research where any researcher, governmental agency or private entity can submit and evaluate their latent fingerprint recognition methods. The platform includes the benchmarking of the latent fingerprint recognition performance against commonly used approaches. The detailed analysis can provide the performance limitations of latent fingerprint recognition mechanisms to researchers and promote them to develop more robust latent fingerprint recognition algorithms.
\end{itemize}

The paper is organized as follows. Section \ref{sec:relatedworks} provides an overview of past studies which are related to our work in latent fingerprint recognition and databases. We introduce the large-scale LFIW database where the details of the whole dataset are illustrated in Section \ref{sec:database}. The evaluated fingerprint recognition algorithms are introduced in Section \ref{sec:algorithms}, followed by a detailed discussion of benchmarking results in Section \ref{sec:results}. Finally,  Section \ref{sec:conclusion} draws the conclusions.  

% In Section \ref{sec:evaluationplatform}, we present the newly designed independent evaluation platform.

\section{Related works}
\label{sec:relatedworks}

\begin{figure}[t]
\centering
  \includegraphics[width=.9\columnwidth]{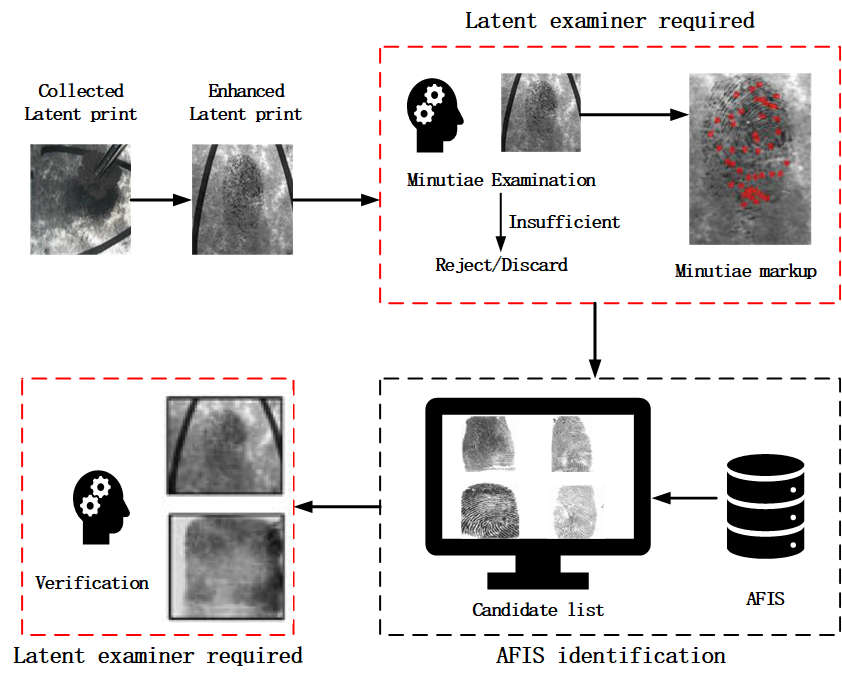}
  \caption{Latent fingerprint recognition with human examiner supervision.}
  \label{fig:latent-recognition}
\end{figure}

Latent fingerprint recognition is a complicated process and the accuracy is generally low. As noted from Fig. \ref{fig:score_drop}, one can note that the comparison scores for the mated samples drop heavily from capacitive to latent fingerprint comparison. Compensation for human examiner supervision (or semi-automatic) can increase the accuracy of latent fingerprint recognition (see Fig. \ref{fig:latent-recognition} for an example of common latent fingerprint recognition workflow). Such a workflow has a significant difference from the common AFIS operation mode (e.g. border control, mobile unlock and payment, etc.). With the rapid development of biometric technology, more and more fully automated latent fingerprint recognition algorithms have been proposed. There are three steps in the automated latent fingerprint recognition system: segmentation, minutiae extraction, and comparison. A brief review of existing approaches for these three steps is given below and the detailed summary is illustrated in Table \ref{tab:sum-recognition}. Prior to discussing each of the components, we also discuss present available datasets for latent fingerprint recognition.

%We present a set of related works grouped under four categroies such as relevant databases, segmentation approaches, minutiae extraction mechanisms and latent comparison approaches. The goal of this section is to present an overview of all works in latent fingerprint recognition before discussing the challenges in large-scale latent fingerprint recognition.

\begin{table*}
\tablestyle[sansboldbw]
\caption{Latent fingerprint databases.}
\resizebox{\textwidth}{!}{
\begin{tabular}{*{8}{p{0.14\textwidth}p{0.08\textwidth}p{0.07\textwidth}p{0.08\textwidth}p{0.12\textwidth}p{0.1\textwidth}p{0.1\textwidth}p{0.1\textwidth}}}
\theadstart
    \thead DBs &
    \thead Unique Subjects &
    \thead Total Size &
    \thead Latent Size &
    \thead Finger photo included &
    \thead Reference included &
    \thead Multiple Surface &
    \thead Availability \\
\tbody
 %
 %\tsubheadstart
 
 NIST SD27 \cite{garris2000nist} & N/A & 258 & 258 & \XSolidBrush & \Checkmark & \XSolidBrush & \XSolidBrush \\
 WVU \cite{wvudatabase} & 35 & 449 & 449 & \XSolidBrush & \Checkmark & \XSolidBrush & \XSolidBrush \\
 MOLF \cite{sankaran2015multisensor} & 100 & 6,000 & 6,000 & \XSolidBrush & \XSolidBrush & \XSolidBrush & \Checkmark \\
 TOLF \cite{feng2012robust} & 10 & 112 & 100 & \XSolidBrush & \XSolidBrush & \XSolidBrush & \Checkmark \\
 IIIT-D \cite{sankaran2011matching} & 15 & 1,046 & 1,046 & \XSolidBrush & \Checkmark & \XSolidBrush & \Checkmark \\
 SLF \cite{sankaran2012hierarchical} & 30 & 1,140 & 1,140 & \XSolidBrush & \Checkmark & \XSolidBrush & \Checkmark \\
 ELFT-EFS \cite{paulino2012latent} & 837 & 1,114 & 1,066 & \XSolidBrush & \XSolidBrush & \XSolidBrush & \XSolidBrush \\
 IIID-MSLFD \cite{sankaran2015latent} & 51 & 1,571 & 1,080 & \XSolidBrush & \Checkmark & \Checkmark & \Checkmark \\
 \textcolor{blue}{\textbf{Proposed - LFIW}} & \textcolor{blue}{\textbf{132}} & \textcolor{blue}{\textbf{13,180}} & \textcolor{blue}{\textbf{9,224}} & \textcolor{blue}{\textbf{\Checkmark}} & \textcolor{blue}{\textbf{\Checkmark}} & \textcolor{blue}{\textbf{\Checkmark}} & \textcolor{blue}{\textbf{\Checkmark$^*$}} \\
 \tend
\end{tabular}
}
\begin{flushleft}
     \textit{\textbf{$^*$}The proposed database will be publicly available upon the acceptance of this paper.}
\end{flushleft}
\label{tab:sum-database}
\end{table*}

\subsection{Latent fingerprint databases}

%In order to learn the performance of developed latent fingerprint recognition algorithms, it is common to evaluate them on some latent fingerprint databases. 
There are several existing latent fingerprint databases available for performance evaluation, such as, West Virginia University (WVU) database \cite{paulino2012latent}, Multisensor Optical and Latent Fingerpr (MOLF) database \cite{sankaran2015multisensor}, Tsinghua Latent Overlapped Fingerprint (TOLF) database \cite{feng2012robust}, and so on. A list of commonly used latent fingerprint databases is given in Table \ref{tab:sum-database}. However, NIST SD27 has been withdrawn and is no longer available.

%Table \ref{tab:sum-recognition} presents the most commonly used latent fingerprint databases \cite{garris2000nist}.

As we can see from Table \ref{tab:sum-database}, the size of existing latent fingerprint databases is small and the simulation of real-world scene latent fingerprints is a very challenging issue. Unfortunately, the only real crime scenes database NIST SD27 has been withdrawn. To develop latent fingerprint recognition techniques, creating a new and challenging database is needed. Therefore, a new database that meets the following requirements is desirable:
\begin{enumerate}
    \item Large-scale database including large number of unique finger instances and not just multiple fingers from a few unique subjects;
    \item Real-world scenarios where latent fingerprints vary in terms of quality, resolution and material surfaces;
    \item Both reference and latent fingerprints are available for each unique fingerprint instance in contact-less and contact-based scenarios;
    %\item In addition to the latent fingerprint images, benchmarking results obtaining from commonly used fingerprint recognition methods should be provided;
    \item Public availability of dataset for academic research under different latent and cross-sensor (latent-vs-contact-based) protocols.
\end{enumerate}

Compared to other existing latent fingerprint databases, the proposed LFIW dataset not only has the largest number of total fingerprint samples (13,180) from various scenarios but also has the largest amount of unique fingerprint instances (1318 from 132 subjects). The database further meets the criteria mentioned above.

\subsection{Latent fingerprint segmentation techniques}

Latent fingerprint segmentation can be defined as the separation of the fingerprint region from the entire image. Segmentation methods with high accuracy can not only reduce the computational complexity but also usually improve the minutiae extraction performance. There are two types of segmentation tasks: separating non-overlapping and overlapping latent fingerprints. For non-overlapped latent fingerprint segmentation task, an extended directional total variation model was developed by Zhang \textit{et al.} \cite{zhang2012latent} to search for and separate latent fingerprints from the background. Cao \textit{et al.} \cite{cao2014segmentation} presented a dictionary based approach to segment latent fingerprints as well as improve their quality. Many machine-deep learning (ML/DL)-based latent fingerprint segmentation approaches have been developed over the past few years. Patches from the region of interest of an image are trained in a convolutional neural network (CNN) and used for segmentation by Stojanovic \textit{et al.} \cite{stojanovic2016fingerprint}. A foreground (latent fingerprint) and background classification method was developed by Sankaran \textit{et al.} \cite{sankaran2017adaptive}, which takes advantage of random decision forest. Nguyen \textit{et al.} \cite{nguyen2018automatic} introduced a CNN-based latent fingerprint segmentation algorithm (SegFinNet) to compensate for the insufficient performance of existing Commercial Off-The-Shelf (COTS) latent fingerprint recognition methods. Compared to non-overlapped latent fingerprint segmentation, separating overlapped fingerprints from each other and from the background is challenging. Chen \textit{et al.} \cite{chen2011separating} applied local Fourier transform and relaxation labelling to segment overlapped fingerprints. To overcome the shortcomings of relaxation labelling-based methods, Zhao and Jain \cite{zhao2012model} developed a zero-pole model, Legendre polynomial, 2D Fourier Expansion, and monomial basis function for overlapped fingerprint segmentation. An adaptive neuro-fuzzy inference system classifier was used for overlapping fingerprint segmentation by Jeyanthi \textit{et al.} \cite{jeyanthi2016efficient}. Stojanovic \textit{et al.} \cite{stojanovic2017novel} combined neural networks and Fourier analysis to separate the overlapping fingerprints.

\subsection{Latent fingerprint minutiae extraction techniques}

Many fingerprint minutiae extraction methods have been developed in the past, however, the number of minutiae extraction algorithms that are especially used for the latent fingerprints is limited. Su and Srihari \cite{su2010latent} developed a latent fingerprint minutiae extraction approach using a regression Gaussian process model to estimate the location of finger core points and orientation fields. Sankaran \textit{et al.} \cite{sankaran2014latent} presented to classify minutia or non-minutia regions in a latent fingerprint by using stacked denoising sparse auto-encoders. Tang \textit{et al.} \cite{tang2017latent} used a fully connected CNN to extract minutiae from the complicated background so that latent fingerprint segmentation and quality enhancement are no longer needed in this approach.

\subsection{Latent fingerprint comparison pipeline}

It is not a simple task to find a match between an unknown fingerprint and a fingerprint in a big database, while this becomes even more difficult for latent fingerprint. Jain and Feng \cite{jain2010latent} combined extended fingerprint features and minutiae to perform latent fingerprint comparison. Paulino \textit{et al.} \cite{paulino2012latent} applied Descriptor-Based Hough Transform (DBHT) to compare reconstructed orientation fields in two latent fingerprints. Cao and Jain \cite{cao2018automated} proposed to generate two minutiae templates (obtained from CNN-based and dictionary-based ridge flow, respectively) and one texture template (virtual minutiae) for latent fingerprint comparison. In addition to use minutiae for matching, pores were also used by Nguyen and Jain \cite{nguyen2019end} to increase the accuracy of latent fingerprint comparison.

%In the following sections, we will first introduce a new large-scale wild latent fingerprint database, as well as a evaluation platform and its benchmarking results.

\begin{table*}
\tablestyle[sansboldbw]
\caption{Latent fingerprint recognition techniques.}
\resizebox{\textwidth}{!}{
\begin{tabular}{*{4}{p{0.16\textwidth}p{0.35\textwidth}p{0.11\textwidth}p{0.27\textwidth}}}
\theadstart
    \thead Reference &
    \thead Technique &
    \thead DB &
    \thead Evaluation \\
\tbody
 \tsubheadstart
 \multicolumn{4}{c}{\textbf{Non-overlapped latent fingerprint segmentation methods}} \\
 Zhang \textit{et al.} \cite{zhang2012latent}  & Extended directional total variation model & NIST SD27 & Subjective inspection \\
 Cao \textit{et al.} \cite{cao2014segmentation}  & Course and fine ridge structure dictionary & SD27\& WVU & R1-IA=0.61(SD27)/0.70(WVU) \\
 Stojanovic \textit{et al.} \cite{stojanovic2016fingerprint}  & CNN-based & FVC2002 \cite{maio2002fvc2002} & EC=0.039(noisy)/0.037(regular) \\
 Sankaran \textit{et al.} \cite{sankaran2017adaptive} & Random decision forest & NIST SD4/27 IIITD-CLF \cite{sankaran2015latent} & R50-IA=0.723 \\
 Nguyen \textit{et al.} \cite{nguyen2018automatic} & Fully convolutional neural network & NIST SD27 WVU \cite{paulino2012latent} & MDR=0.026(SD27)/0.131(WVU), FDR=0.164(SD27)/0.053(WVU) \\
 \tsubheadstart
 \multicolumn{4}{c}{\textbf{Overlapped latent fingerprint segmentation methods}} \\
 Chen \textit{et al.} \cite{chen2011separating}  & Fourier transform and relaxation labelling & FVC2002 \cite{maio2002fvc2002} & Insufficient for singular points \\
 % Feng \textit{et al.} \cite{feng2012robust}  & Constrained relaxation labeling & FVC2002 \cite{maio2002fvc2002} & Insufficient for singular points \\
 Zhao and Jain \cite{zhao2012model}  & Joint orientation modeling & NIST SD27 FVC2002 \cite{maio2002fvc2002} & Better manual marking minutiae \\
 Jeyanthi \textit{et al.} \cite{jeyanthi2016efficient}  & Adaptive neuro fuzzy inference system & NIST SD27 FVC2006 \cite{cappelli2007fingerprint} & SR=0.924, CR=0.886, AC=0.92 \\
 Stojanovic \textit{et al.} \cite{stojanovic2017novel}  & Neural networks and Fourier analysis & TOLF \cite{feng2012robust} & R1-IA=0.94(S)/0.91(M)/0.90(L) \\
 %
%  \tsubheadstart
%  \multicolumn{4}{c}{\textbf{Latent fingerprint enhancement methods}} \\
%  %
%  Yoon \textit{et al.} \cite{yoon2010latent,yoon2011latent}  & STFT/R-RANSAC orientation
% estimation & NIST SD27 &  Better matching accuracy \\
%  Feng \textit{et al.} \cite{feng2012orientation} & Orientation patches dictionary & NIST SD27 & AEE=18.44 \\
%  Kumar \textit{et al.} \cite{kumar2015latent}  & Look-up table of orientation patches & NIST SD27 & AC=0.8423 \\
%  Cao \textit{et al.} \cite{cao2015latent}  & Orientation estimation using CNN & NIST SD27 & Bettter R1-IA \\
%  Li \textit{et al.} \cite{li2018deep}  & FingerNet noise removal & NIST SD27 & Fast speed and better accuracy \\
%
 \tsubheadstart
 \multicolumn{4}{c}{\textbf{Latent fingerprint minutiae extraction methods}} \\
 Su \textit{et al.} \cite{su2010latent}  & Regression Gaussian process model & NIST SD27/4 &  AC=0.845 \\
 Sankaran \textit{et al.} \cite{sankaran2014latent} & Stacked denoising sparse autoencoders & NIST SD27 & R10-IA=0.3361 \\
 Tang \textit{et al.} \cite{tang2017latent}  & Fully CNN & NIST SD27 & AC=0.534 \\
 \tsubheadstart
 \multicolumn{4}{c}{\textbf{Latent fingerprint comparison methods}} \\
 Jain \textit{et al.} \cite{jain2010latent}  & Extended fingerprint features and minutiae & NIST SD27/4 &  R1-IA=0.74 \\
 Paulino \textit{et al.} \cite{paulino2012latent} & DBHT & SD27\& WVU & R1-IA is better then VeriFinger SDK \\
 Cao \textit{et al.} \cite{cao2018automated}  & 2 minutiae and 1 texture templates & SD27\& WVU & R1-IA=0.647(SD27)/0.753(WVU) \\
 Nguyen \textit{et al.} \cite{nguyen2019end}  & Using both minutiae and pores & NIST SD30 & R1-IA=0.964(Min)/0.981(M+P) \\
 \tend
\end{tabular}
}
\begin{flushleft}%\scriptsize
     \textit{Note: \textbf{FAR} represents the False Acceptance Rate, \textbf{TAR} represents the True Acceptance Rate, \textbf{MDR} represents the Missed Detection Rate, \textbf{FDR} represents the False Detection Rate, \textbf{R1-IA} represents the Rand 1 Identification Accuracy, \textbf{EC} represents the Error Coefficient, \textbf{R50-IA} represents the Rank-50 identification accuracy, \textbf{SR} represents the Separation Rate, \textbf{CR} represents the Classification Rate, \textbf{AC} represents the accuracy, \textbf{S} represents small, \textbf{M} represents medium, \textbf{L} represents large, \textbf{AEE} represents the Average Estimation Error (in degrees), \textbf{R10-IA} represents the Rank-10 identification accuracy, \textbf{Min} represents the minutiae, and \textbf{M+P} represents the minutiae plus pores.}
\end{flushleft}
\label{tab:sum-recognition}
\end{table*}

\section{Latent Fingerprint in the Wild Database}
\label{sec:database}

As noted in the previous studies, the performance evaluation of existing latent fingerprint recognition techniques is mainly based on only one or certain databases, which are usually limited in size, diversity of image acquisition devices, image quality, and realistic capture environment. The best way to evaluate the performance of a latent fingerprint recognition algorithm is to challenge it using different databases, image acquisition and testing protocols. In order to overcome these limitations and provide a new database for performance evaluation under real-world scenarios with high image quality, the LFIW database created in this work consists of six subsets of which two subsets are traditional fingerprints, three latent fingerprints and one fingerphoto set as provided below:
\begin{enumerate}
    \item \textbf{\textit{R-opt}:} Reference fingerprints from optical sensor;
    \item \textbf{\textit{R-cap}:} Reference fingerprints from capacitive sensor;
    \item \textbf{\textit{Smt}:} Smartphone fingerphotos;    
    \item \textbf{\textit{L-wall}:} Latent fingerprints captured from wall surface;
    \item \textbf{\textit{L-ipad}:} Latent fingerprints captured from Ipad surface;
    \item \textbf{\textit{L-alum}:} Latent fingerprints captured from aluminum foil surface.
\end{enumerate}

Detailed information of the LFIW database is further provided in Table \ref{tab:property-database}. All fingerprint images have been cropped and rotated to remove the background in order to avoid unnecessary variables and facilitate the following processing steps (e.g. enhancement, minutiae extraction, etc.). Examples of the R-opt, R-cap, Smt, L-wall, L-ipad, and L-alum images are illustrated in Fig. \ref{fig:fingerprint_examples} red dotted block.

\subsection{Reference fingerprint images: R-opt and R-cap}

For each of the 132 subjects in the LFIW database, two enrolment images were captured by using two professional fingerprint acquisition sensors: one optical fingerprint sensor and one capacitive sensor. The optical sensor is ZKTeco Live10R fingerprint capture device and the capacitive sensor is Bingup FPW-A360 fingerprint capture device. The original size of the fingerprint images from the optical sensor is $288\times375$ pixels (106 KB) and is $256\times360$ pixels (91 KB) for the capacitive sensor. All reference fingerprint images are in 500 ppi. There are a total of $(132\quad subjects\times10\quad fingers-2\quad lost)\times2\quad sensors\times2\quad enrolment=5272$ reference fingerprint images in the LFIW database. 

% \subsection{Wet fingerprint images: W-opt and W-cap}

% In addition to the reference fingerprints, we also collected wet fingerprints by using optical and capacitive sensors. Subjects were required to get their fingers wet by touch a wet sponge before each acquisition. The other setups are the same as the reference fingerprints acquisition. There are a total of $(132\quad subjects\times10\quad fingers-2\quad lost)\times2\quad sensors\times=2636$ wet fingerprint images in the 'Latent in the Wild' database.
  
\subsection{Smartphone fingerphoto images: Smt}

All smartphone fingerphoto images were taken by Huawei Honor20 smartphone (48+8+2 megapixel triple camera). All subjects were asked to place each of their ten fingers on a white background under additional white light source. The acquisition distance to the fingers and the focus were controlled and the build-in flash has been turned off.  The original size of the fingerprint images is $3000\times4000$ pixels ($\sim$2MB) and the ppi is 96 by default. There are a total of $132\quad subjects\times10\quad fingers-2\quad lost=1318$ smartphone fingerprint images in the LFIW database.

\subsection{Wall surface latent fingerprint images: L-wall}

In order to simulate the latent fingerprints captured on the wall in a real crime scene (indoor environment, such as office, bank, school, etc.), subjects were required to touch all 10 fingers on an office desk partition wall to leave their fingerprints on the wall. Copper powder was used to make fingerprints visible and wall latent fingerprint images were taken by Iphone 8 plus smartphone (12 megapixel dual camera). Additional white light source was used and the acquisition distance to the fingerprints was controlled while the build-in flash has been turned off. The original size of the wall latent fingerprint images is $3024\times4032$ pixels ($\sim$2.5MB) and the ppi is 72 by default. There are a total of $132\quad subjects\times10\quad fingers-2\quad lost=1318$ wall surface latent fingerprint images in the LFIW database.

\subsection{Ipad surface latent fingerprint images: L-ipad}

In order to simulate the latent fingerprints captured on the surface of electronic devices as well as on the glasses in a real crime scene, subjects were required to touch all 10 fingers on an Ipad screen surface (without protective film) to leave their fingerprints on the Ipad screen. Copper powder was used to make fingerprints visible and latent fingerprint images were taken by the same Iphone 8 plus smartphone. The acquisition setups and images properties are the same as the L-wall. Since additional white light source was used, the screen reflection was avoided as much as possible during the acquisition process. There are a total of $132\quad subjects\times10\quad fingers-2\quad lost=1318$ Ipad screen surface latent fingerprint images in the LFIW database.

\subsection{Aluminum foil surface latent fingerprint images: L-alum}

In order to simulate the latent fingerprints captured on the (deformable) metal surface in a crime scene, subjects were required to touch all 10 fingers on an aluminum foil surface to leave their fingerprints on the foil. Copper powder was again used to make fingerprints visible and aluminum foil surface latent fingerprint images were taken by the same Iphone 8 plus smartphone. The acquisition setups and images properties are the same as the L-wall. Since additional white light source was used, the aluminum foil reflection was avoided as much as possible during the acquisition process. There are a total of $132\quad subjects\times10\quad fingers-2\quad lost=1318$ aluminum foil surface latent fingerprint images in the LFIW database.

\subsection{Fingerprint images preprocessing}

All Smt, L-wall, L-ipad, and L-alum original images have been cropped and rotated manually for further processing. Moreover, all preprocessed fingerprint images (JPEG format) have a 500 ppi version, a gray-scale 500 ppi version, a gray-scale 500 ppi PNG format version, and a gray-scale 500 ppi PGM format version.

\begin{table}
\tablestyle[sansboldbw]
\caption{Properties of the 'Latent Fingerprint in the Wild' database.}
\begin{tabular}{*{2}{|p{0.32\columnwidth}|p{0.6\columnwidth}}}
\theadstart
    \thead Property &
    \thead Information \\
\tbody

 Number of subjects  & \textbf{132} \\
 Finger instances per subject & \textbf{5} fingers per hand $\times$ \textbf{2} hands \\
 Gender & \textbf{74} Males + \textbf{58} Females \\
 Age groups & 18-35: \textbf{106}, 36-55: \textbf{19}, 56-70: \textbf{7} \\
 Number of images & \textbf{100} per subject $\times$ \textbf{132} $=$ \textbf{13180} $^{*}$ \\
 \tend
\end{tabular}
\begin{flushleft}
     \textit{$^{*}$ one of the subjects (worker) lost two Thumb fingers.}
\end{flushleft}
\label{tab:property-database}
\end{table}

\section{Fingerprint recognition algorithms}
\label{sec:algorithms}

As described in the previous sections, a number of existing state-of-the-art fingerprint recognition algorithms are evaluated on the new LFIW database. Meanwhile, different versions of minutiae/features have been generated by these evaluated algorithms and stored in the benchmark databases for further evaluation. In this section, we briefly discuss the algorithms that were tested on the LFIW database.

\subsubsection{NIST Biometric Image Software (NBIS) \cite{nbis}}

The NBIS is one of the most well-known fingerprint recognition toolkits that can be freely used and distributed. Two components are used for the performance evaluation: MINDTCT and BOZORTH3. MINDTCT is a minutiae detector and it can automatically locate and record ridge endings and bifurcations in a fingerprint image. BOZORTH3 is a fingerprint comparison algorithm and it is minutiae-based. %It can do both one-to-one and one-to-many matching operations. 
It accepts minutiae generated by the MINDTCT algorithm. All extracted minutiae from MINDTCT are stored in the benchmark databases and can be used for other fingerprint-comparison algorithms in case needed.

\subsubsection{Minutia Cylinder-Code (MCC) fingerprint recognition SDK \cite{mcc,cappelli2010minutia}}

The so-called 'cylinder' is a 3D data structure containing minutiae distances and angles. Any standardized minutiae position and direction (e.g. ISO/IEC 19794-2 \cite{iso2011iec}) can be used as mandatory pre-condition to establish the cylinder. Instead of designing complex metrics to calculate local similarities and generate the comparison score, a very simple algorithm is applied in MCC by taking advantages of the cylinder invariance. MCC uses ISO/IEC minutiae information to generate its own minutiae template for fingerprint comparison \cite{dorizzi2009fingerprint,biolab}. %All minutiae templates generated by MCC are also stored in the benchmark databases and can be used for other fingerprint matching algorithms in case needed.

\subsubsection{VeriFinger fingerprint recognition SDK \cite{verifinger}}

VeriFinger is a commercial fingerprint recognition software designed for biometric systems developers and integrators by Neurotechnology \cite{verifinger}. The software can conduct fast fingerprint comparison in 1-to-1 and 1-to-many modes. The VeriFinger algorithm is based on deep neural networks and follows the commonly accepted fingerprint recognition scheme, which uses a set of minutiae along with a number of proprietary algorithmic solutions that enhance system performance and reliability. VeriFinger can produce its own minutiae template for fingerprint comparison. It has also been submitted to the FVC-onGoing \cite{dorizzi2009fingerprint,biolab} framework and has reached NIST MINEX compliance. %All VeriFinger mimutiae templates are stored in the benchmark databases and can be used for other fingerprint matching algorithms in case needed.

\subsubsection{MinutiaeNet minutiae extractor \cite{Nguyen_MinutiaeNet}}

MinutiaeNet can perform fully automatic latent fingerprint minutiae extraction by using two independent deep neural networks. The first network is named as CoarseNet and it estimates the minutiae score map and minutiae orientation based on CNN and fingerprint domain knowledge (enhanced image, orientation field, and segmentation map). FineNet is the second network and it refines the candidate minutiae locations based on the score map. MinutiaeNet has been particularly tested on NIST SD27 latent fingerprint database and the performance is better than several other state-of-the-art minutiae extraction algorithms. However, MinutiaeNet needs to apply other methods for minutiae comparison, such as MCC or BOZORTH3. %All minutiae files produced from MinutiaeNet are stored in the benchmark databases and can be used for other fingerprint matching algorithms in case needed. 

\subsubsection{MSU Latent Automatic Fingerprint Identification System \cite{cao2019end}}

MSU-LAFIS is an end-to-end latent fingerprint search system, which has five main steps: 1) fingerprint region of interest segmentation, 2) segmented image pre-processing, 3) feature extraction, 4) feature comparison, and 5) comparison results generation. Two isolated feature extraction algorithms are used to produce additional feature templates. In order to avoid an insufficient number of extracted features from latent fingerprints (too small area or very low image quality), the feature template can be established by combining real extracted features and a group of generated virtual features. Each latent fingerprint feature and its neighbourhood are used to obtain a 96-dimensional descriptor for feature comparison. The descriptor length of the virtual feature is further compressed from 96 to 16 to increase processing speed by using product quantization. %All MSU-LAFIS feature templates are stored in the benchmark databases and can be used for other fingerprint matching algorithms in case needed.

\section{Protocols, Results and Discussion}
\label{sec:results}

\begin{table*}[t]
\tablestyle[sansboldbw]
\caption{Performance indicators measured on the LFIW database for the overall comparison experiments.}
\begin{tabular}{*{11}{|p{0.08\textwidth}|p{0.05\textwidth}|p{0.05\textwidth}|p{0.05\textwidth}|p{0.04\textwidth}|p{0.06\textwidth}|p{0.06\textwidth}|p{0.08\textwidth}|p{0.09\textwidth}|p{0.08\textwidth}|p{0.08\textwidth}}}
\theadstart
    \thead Algorithm &
    \thead GMean &
    \thead GSTD &
    \thead IMean &
    \thead ISTD &
    \thead AUC(\%) &
    \thead EER(\%) &
    \thead ZFMR(\%) &
    \thead FMR100(\%) &
    \thead FMR20(\%) &
    \thead FMR10(\%) \\
\tbody
 %
 %\tsubheadstart
 
 NBIS  & 15.21 & 32.13 & 4.22 & 3.59 & 57.91 & 46.10 & 90.79 & 80.58 & 75.25 & 70.94 \\
 MCC & 0.03 & 0.04 & 0.01 & 0.002 & 75.44 & 32.27 & 99.82 & 63.60 & 56.77 & 51.17 \\
 VeriFinger & 162.66 & 128.21 & 6.79 & 5.73 & 85.44 & 22.82 & 55.08 & 30.78 & 29.12 & 27.53 \\
 MN-MCC & -0.15 & 0.37 & 0.01 & 0.002 & 46.36 & 51.85 & 99.43 & 94.31 & 89.24 & 84.55 \\
 MN-NBIS & 1.30 & 5.11 & 0.34 & 1.04 & 54.77 & 45.61 & 98.56 & 91.85 & 89.10 & 81.29 \\
 MSU & 28.87 & 56.75 & 3.55 & 5.34 & 57.10 & 47.51 & 100.00 & 75.38 & 72.82 & 70.66 \\
 \tend
\end{tabular}
\begin{flushleft}\scriptsize
     \textit{Note: \textbf{GMean:} Mated scores distribution mean; \textbf{GSTD:} Mated scores distribution standard deviation; \textbf{IMean:} Impostor scores distribution mean; \textbf{ISTD:} Non-mated scores distribution standard deviation; \textbf{AUC:} Area under the ROC curve; \textbf{EER:} Equal Error Rate; \textbf{FMR100/20/10:} False match rate when false non match rate is 1\%/20\%/10\%.}
\end{flushleft}
\label{tab:all-errors-results}
\end{table*}

With the newly introduced dataset, we also conduct an extensive evaluation by introducing three different protocols. The first protocol is to establish the baseline performance in traditional fingerprint capture devices (optical and capacitive sensors). The second protocol is to evaluate the scenarios of comparing the latent-vs-latent and latent fingerprint with traditional contact-based fingerprints. The third protocol is to account for comparison of latent fingerprints with contactless fingerprints derived from fingerphotos. With our protocols, we cover all possible scenarios of relevance in real-world use cases. Given the large scale of LFIW dataset, we also perform both verification and identification experiments to provide the reader with an understanding of the challenges and thereby suggest directions for future works. 

\subsection{Verification results - Overall}
\begin{figure}[]
\centering
\includegraphics[width=.8\columnwidth]{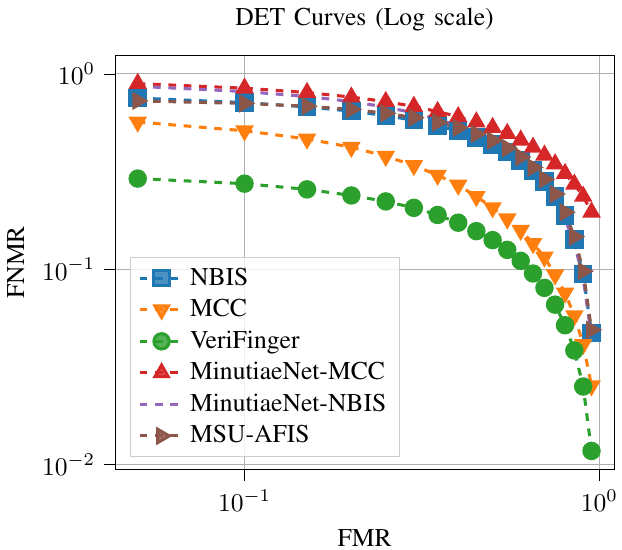}
 \caption{DET curve of the overall comparison experiments for the LFIW database.}
 \label{fig:all-det-results}
\end{figure}

Before each of the protocols is considered, we provide an overall evaluation of the dataset by combining all the images of LFIW dataset. In the overall evaluation experiment, a total of 118,620 mated comparison scores and 173,593,780 non-mated comparison scores are generated. The Detection Error Tradeoff (DET) curves of the overall comparison experiments for the LFIW dataset are presented in Fig. \ref{fig:all-det-results} along with the detailed results for various metrics in Table \ref{tab:all-errors-results}. Two algorithms, VeriFinger and MCC perform slightly better than the average, however, the Equal Error Rate (EER) is 22.82\% and 32.27\% respectively. Even the MinutiaeNet and MSU-LAFIS which are particularly designed for latent fingerprint recognition perform poorly with an EER of 51.85\% (MinutiaeNet-MCC)/45.61\% (MinutiaeNet-NBIS) and 47.51\%, respectively. We have therefore analyzed the causes of low performance and observe a high Failure To Enrol Rate (FTER) for most of the selected algorithms. The overall FTER are: $FTER_{NBIS}=25.2\%, FTER_{MCC}=36.68\%, FTER_{VeriFinger}=40.78\%, FTER_{MN-MCC}=28.1\%, FTER_{MN-NBIS}=26.01\%, FTER_{MSU-AFIS}=7.67\%$, and FTER for different protocols are given in Table \ref{tab:5-errors-results} in blue color. Such a high FTER for the second and third protocol in Table \ref{tab:5-errors-results} indicates the difficulty to extract minutiae from images in LFIW making it a challenging dataset.

Although some minutiae (features) can be successfully extracted, the number of high quality minutiae (features) might be insufficient for an eligible comparison. MSU-LAFIS has the lowest FTER, NBIS and MinutiaeNet (both MCC and NBIS) have the FTER lower than $30\%$. Since MSU-LAFIS is especially developed for latent fingerprint, so it corresponds to our expectation that it can handle more than $90\%$ of the latent fingerprints in the LFIW database. The FTER for VeriFinger reaches more than $40\%$ which means that almost half of the latent fingerprints in the LFIW database cannot be used for recognition or identification when using VeriFinger algorithm. On one hand, the high FTER from selected non-latent-oriented methods indicates that the latent fingerprints in the LFIW database are much more difficult to be processed compared to existing fingerprints and latent fingerprints. Therefore, robust latent fingerprint recognition algorithms are needed. On the other hand, the pre-processing approaches from the selected algorithms can be optimized to be able to handle latent fingerprints in the LFIW database.

% \begin{figure*}[t]
% \centering
%   \centering
%      \begin{subfigure}[b]{0.32\columnwidth}
%          \centering
%        \includegraphics[width=\textwidth]{fig5-1.pdf}
%          \caption{NBIS}
%      \end{subfigure}
%      \hfill
%      \begin{subfigure}[b]{0.32\columnwidth}
%          \centering
%         \includegraphics[width=\textwidth]{fig5-2.pdf}
%          \caption{MCC}
%      \end{subfigure}
%      \hfill
%      \begin{subfigure}[b]{0.32\columnwidth}
%          \centering
%         \includegraphics[width=\textwidth]{fig5-3.pdf}
%          \caption{VeriFinger}
%      \end{subfigure}
%      \hfill
%      \begin{subfigure}[b]{0.32\columnwidth}
%          \centering
%         \includegraphics[width=\textwidth]{fig5-4.pdf}
%          \caption{MN-MCC}
%      \end{subfigure}
%      \hfill
%      \begin{subfigure}[b]{0.32\columnwidth}
%          \centering
%         \includegraphics[width=\textwidth]{fig5-5.pdf}
%          \caption{MN-NBIS}
%      \end{subfigure}
%      \hfill
%      \begin{subfigure}[b]{0.32\columnwidth}
%          \centering
%         \includegraphics[width=\textwidth]{fig5-6.pdf}
%          \caption{MSU-LAFIS}
%      \end{subfigure}

% \begin{subfigure}[b]{0.5\columnwidth}
% \centering
%     \includegraphics[width=0.5\columnwidth]{fig4_legend.pdf}
%     \caption*{Legend}
% \end{subfigure}    
%         \caption{Distributions of the overall comparison scores in the LFIW database (Enlarged version in Appendix under Fig~\ref{fig:appendix-all-distribution-results})}
%         \label{fig:all-distribution-results}
% \end{figure*}

\begin{figure*}[htp]
\centering
  \centering
     \begin{subfigure}[b]{0.32\textwidth}
         \centering
       \includegraphics[width=\textwidth]{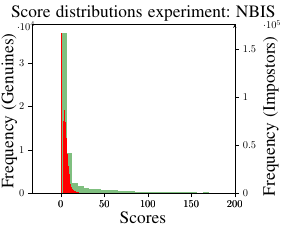}
         \caption{NBIS}
     \end{subfigure}
     \hfill
     \begin{subfigure}[b]{0.32\textwidth}
         \centering
        \includegraphics[width=\textwidth]{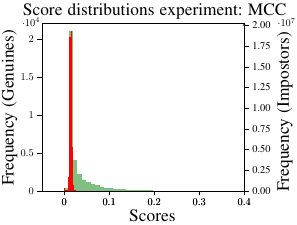}
         \caption{MCC}
     \end{subfigure}
     \hfill
     \begin{subfigure}[b]{0.32\textwidth}
         \centering
        \includegraphics[width=\textwidth]{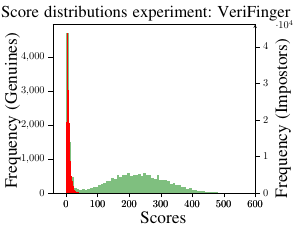}
         \caption{VeriFinger}
     \end{subfigure}
     \\
     \begin{subfigure}[b]{0.32\textwidth}
         \centering
        \includegraphics[width=\textwidth]{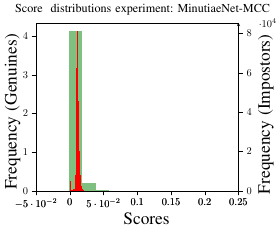}
         \caption{MN-MCC}
     \end{subfigure}
     \hfill
     \begin{subfigure}[b]{0.32\textwidth}
         \centering
        \includegraphics[width=\textwidth]{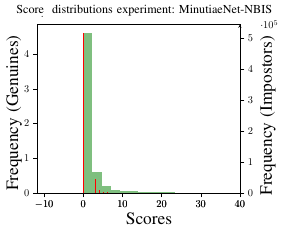}
         \caption{MN-NBIS}
     \end{subfigure}
     \hfill
     \begin{subfigure}[b]{0.32\textwidth}
         \centering
        \includegraphics[width=\textwidth]{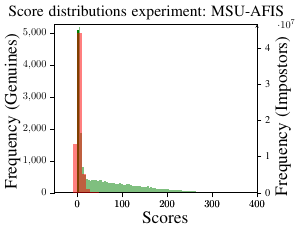}
         \caption{MSU-LAFIS}
     \end{subfigure}

\begin{subfigure}[b]{0.5\columnwidth}
\centering
    \includegraphics[width=0.5\columnwidth]{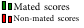}
    \caption*{Legend}
\end{subfigure}    
        \caption{Distributions of the overall comparison scores in the LFIW database}
        \label{fig:all-distribution-results}
\end{figure*}

The distributions of the overall comparison scores from the selected algorithms are illustrated in Fig. \ref{fig:all-distribution-results}. It can be observed that none of the selected algorithms can well separate the mated scores and the non-mated scores well. The highest frequency mated and non-mated scores are almost overlapping for all the methods. Compared with other methods, the mated scores of VeriFinger are more distributed far away from the non-mated scores (see Fig. \ref{fig:all-distribution-results} (c)). This is probably due to the high FTER where latent fingerprints have been rejected during the enrollment phase while it could process reference fingerprints captured by using optical and capacitive sensors. There are also a number of mated comparison scores separated far from the non-mated scores for MSU-LAFIS in Fig. \ref{fig:all-distribution-results} (f), but the proportion is lower than VeriFinger. The mated and non-mated scores of MinutiaeNet (both MCC and NBIS) are highly overlapped. In addition to the distributions of the comparison scores, we present the most important performance indicators measured on the LFIW database for the overall comparison experiments in Table \ref{tab:all-errors-results}. It can be noted that VeriFinger has the highest Area Under the ROC Curve (AUC) value and lowest EER. Except VeriFinger, MCC out perform the rest of the algorithms, however, its AUC and EER is still far from indicating the properties of a robust latent fingerprint recognition system. Nevertheless, it is difficult to distinguish whether those higher mated comparison scores are from reference comparisons or latent comparisons by only looking at the overall experimental results. Therefore, in the following parts we will investigate the results for reference and latent comparisons, respectively. %The ROC curve, FMR and FNMR curves, and the full performance indicators are illustrated in Section A of the Supplementary Material (Fig. 13-15, and Table XII).   

\begin{figure*}[]
\centering
\begin{minipage}[b]{0.78\textwidth}
  \centering
     \begin{subfigure}[b]{0.32\textwidth}
         \centering
        \includegraphics[width=\textwidth]{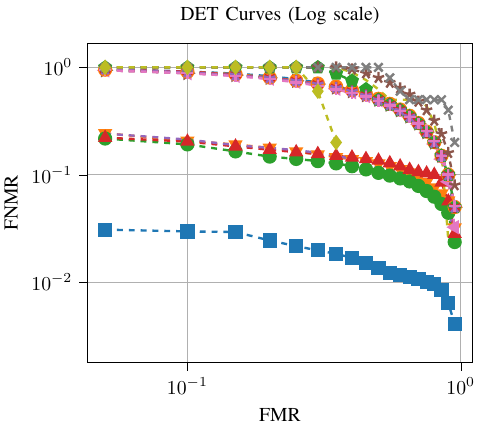}
         \caption{NBIS}
     \end{subfigure}
     \hfill
     \begin{subfigure}[b]{0.32\textwidth}
         \centering
         \includegraphics[width=\textwidth]{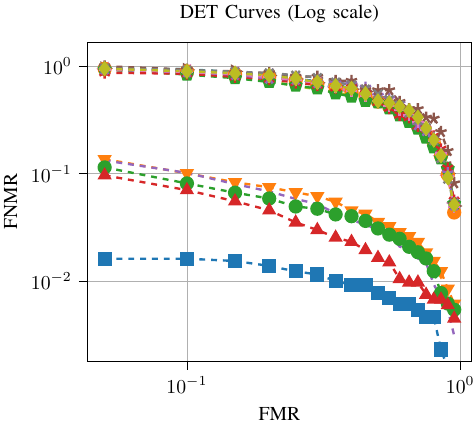}
         \caption{MCC}
     \end{subfigure}
     \hfill
     \begin{subfigure}[b]{0.32\textwidth}
         \centering
         \includegraphics[width=\textwidth]{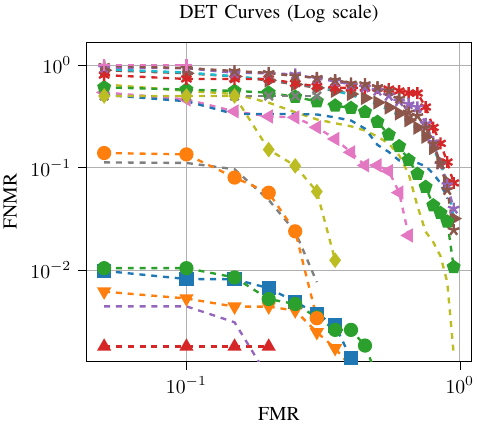}
         \caption{VeriFinger}
     \end{subfigure}
     \hfill
     \begin{subfigure}[b]{0.32\textwidth}
         \centering
        \includegraphics[width=\textwidth]{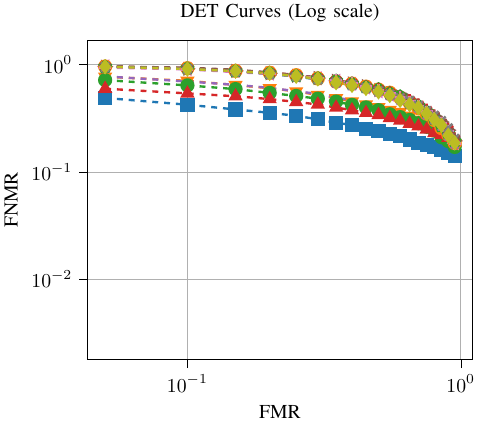}
         \caption{MN-MCC}
     \end{subfigure}
     \hfill
     \begin{subfigure}[b]{0.32\textwidth}
         \centering
         \includegraphics[width=\textwidth]{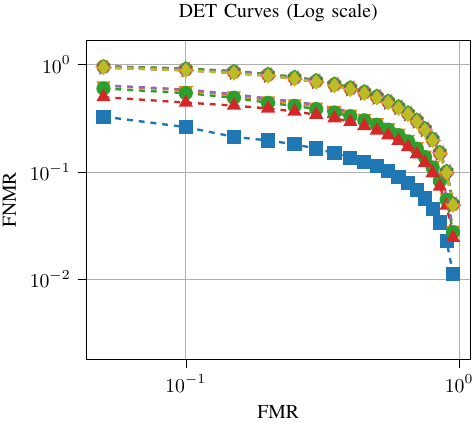}
         \caption{MN-NBIS}
     \end{subfigure}
     \hfill
     \begin{subfigure}[b]{0.32\textwidth}
         \centering
         \includegraphics[width=\textwidth]{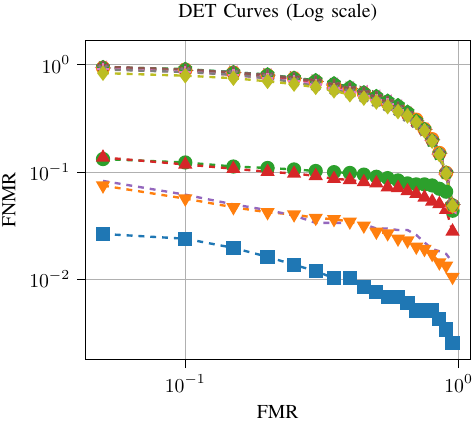}
         \caption{MSU-LAFIS}
     \end{subfigure}
\end{minipage}
\begin{minipage}[b]{0.21\textwidth}
        \includegraphics[width=\textwidth]{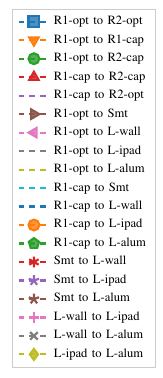}
        \caption*{Legend}
\end{minipage}    
        \caption{DET curves of the NBIS, MCC, VeriFinger, MinutiaeNet-MCC, MinutiaeNet-NBIS, and MSU-LAFIS comparison experiments for the LFIW database}
        \label{fig:5-det-results}
\end{figure*}

\subsection{Protocol I: Verification results - Traditional Sensors}

We illustrate the DET curves of the NBIS, MCC, VeriFinger, MinutiaeNet-MCC, MinutiaeNet-NBIS, and MSU-LAFIS comparison experiments for the LFIW database in Fig. \ref{fig:5-det-results} and the most important performance indicators measured for the selected algorithms in Table \ref{tab:5-errors-results}. From Fig. \ref{fig:5-det-results} and Table \ref{tab:5-errors-results} we can observe that the reference comparisons (e.g. R1-opt to R2-opt, R2-opt to R2-cap) have better performance than latent comparisons (e.g. R2-opt to L-wall, L-ipad to L-alum). Except for VeriFinger, the best performance in reference comparisons is from 'R1-opt to R2-opt (reference fingerprints from optical sensor session 1 \textbf{\textit{vs.}} reference fingerprints from optical sensor session 2)' (see blue dashed lines with square markers in Fig. \ref{fig:5-det-results}). The comparisons 'R1-cap to R2-cap' give the best performance for VeriFinger. Although the comparisons 'R1-cap to R2-cap' (red dashed lines with triangle markers) are between the same acquisition device, the performance is lower than 'R1-opt to R2-opt' for most of the algorithms. It means that the utility of reference fingerprints from optical sensor is better than the capacitive sensor measured by NBIS, MCC, MinutiaeNet (both MCC and NBIS), and MSU-LAFIS. The performance of the remaining reference comparisons are very similar. In Table \ref{tab:5-errors-results}, the EER from the 'R1-cap to R2-cap' comparison experiment is 0.82\% for MCC, which is also higher than the 'R1-opt to R2-opt' experiment. However, both the above mentioned two EERs are smaller (0.11\% and 0.89\%, respectively) than the NBIS.  Moreover, the difference between 'R1-opt to R2-opt' and 'R1-cap to R2-cap' comparison experiments for MCC are also 0.78 less, compared to NBIS. It means that MCC has a better ability to process the reference fingerprints in the LFIW database than NBIS. We can see from Table \ref{tab:5-errors-results} that the overall EERs and FMR100s for VeriFinger are lower than NBIS and MCC, which is the same as we already discussed previously in the overall results section. By comparing Fig. \ref{fig:5-det-results} (d) and (e), as well as EER values for MinutiaeNet we can discover that, NBIS has slightly better overall system performance than MCC when using the extracted minutiae from MinutiaeNet. However, neither MCC nor NBIS can achieve better system performance for MinutiaeNet compared to other fingerprint recognition systems.
 
\subsubsection{Protocol Ia: Verification results - Traditional Cross-Sensors}
We also study the cross-sensor recognition for the completeness of the analysis by comparing the optical v/s capacitive sensors. However, the recognition performance of  'R1-opt to R1-cap' (orange dashed line with triangle markers) and 'R1-cap to R2-opt' (purple dashed line) are lower than same-sensor comaprison for MSU-LAFIS (see Fig. \ref{fig:5-det-results} (f)).

\subsection{Protocol II: Verification results - Traditional v/s Latent}

We further consider a realistic evaluation scenrio where the latents are to be compared against traditional fingerprints. As noted from  Table \ref{tab:5-errors-results}, the overall performance for latent v/s traditional fingerprints is low. The EER values are around 50\% and the FMR100 values are close to 100\% for most of the reference to latent and latent to latent comparisons from all algorithms. The results suggest that comparing latent fingerprints in the LFIW database is a very complex task for the selected algorithms. An interesting EER value 18.8\% can be observed from the 'L-ipad to L-alum' comparison experiment for NBIS. This EER value is much less than the others obtained from latent fingerprints comparisons. Introspecting the comparison scores, we note a very high FTER in NBIS for 'L-ipad to L-alum' comparisons resulting in a misleading low EER. The results also indicate that latent fingerprints captured from Ipad surface and from aluminum foil surface are the most difficult ones for NBIS to extract minutiae. We can see the EER values for VeriFinger are low for many reference fingerprints to latent fingerprints comparison (e.g. EER for R1-opt to L-ipad is 10.2\% and latent fingerprints to latent fingerprints comparison (e.g. EER for L-wall to L-ipad is 7.1\%) experiments. After investigating the comparison scores from these comparison experiments, we explore that the number of scores is quite small. For example, there are 29 mated scores left for L-wall to L-ipad comparison, there is only one mated score left for L-wall to L-alum comparison experiments. All the above discovered atypical EER and FMR values (e.g. low for NBIS and high for VeriFinger) are due to the high FTER that already discussed previously (noted in Table \ref{tab:5-errors-results}). Although MinutiaeNet has been tested on NIST SD27 latent fingerprint database and MSU-LAFIS is especially developed for latent fingerprints, they still fail to provide robust latent fingerprints recognition performance on the LFIW database after looking at the EER and FMR100 values in Table \ref{tab:5-errors-results}.

\begin{table*}[htbp]
\centering
\caption{Performance indicators measured on the LFIW database for the six different algorithms.}
\resizebox{0.99\textwidth}{!}{
	\begin{tabular}{lcccccc}
		\hline
        \multicolumn{7}{c}{\textbf{Fingerprint Recognition Algorithm}} \bigstrut\\
        \cline{2-7} & \textbf{NBIS} &  \textbf{MCC} & \textbf{VeriFinger} & \textbf{MN-MCC} & \textbf{MN-NBIS}  & \textbf{MSU-LAFIS} \bigstrut\\
        \hline
		\multirow{2}[4]{*}{Protocol} & FTER / EER   & FTER  / EER   & FTER  / EER   & FTER  / EER   & FTER  / EER   & FTER  / EER \bigstrut\\
		& GEER  / FMR100 & GEER  / FMR100 & GEER  / FMR100 & GEER  / FMR100 & GEER  / FMR100 & GEER / FMR100 \bigstrut\\
		\hline
		\multicolumn{6}{c}{\textbf{Reference to reference}} \bigstrut\\
		\hline
		\multirow{2}[4]{*}{ R1-opt to R2-opt} &  0.21   /  3.34  &  0.28 /    2.21  &  0.50  /  1.20  &  0.19 /    30.66 &  0.24 /    18.14 &  0.00 /    3.29  \bigstrut\\
		&  3.83  /  4.53     &   2.49 /    2.52     &  1.32   /  1.21     &  32.57   /  63.26    &   16.67       /  45.73    &   3.29       /  4.00     \bigstrut\\
		\hline
		\multirow{2}[4]{*}{ R1-opt to R1-cap} &  0.12   /  17.65 &  0.20 /    10.00 &  0.33 /    0.82  &  0.07      /  43.48 &  0.15 /    32.91 &  0.00  /   6.83  \bigstrut\\
		&  17.88  /  29.62    &   11.33       /  20.00    &   0.79       /  0.95     &   40.54 /    87.03    &   35.89 /    72.60    &   6.83       /  10.96    \bigstrut\\
		\hline
		\multirow{2}[4]{*}{ R1-opt to R2-cap} &  0.12   /  15.00 &  0.19 /    8.53  &  0.32  /  1.02  &  0.08 /   41.56 &  0.15 /    32.05 &  0.00 /   11.93 \bigstrut\\
		&  16.66  /  26.39    &   8.26       /  17.48    &   1.06       /  1.01     &   44.61       /  82.45    &   33.94 /    72.22    &   11.93       /  16.75    \bigstrut\\
		\hline
		\multirow{2}[4]{*}{ R1-cap to R2-cap} &  0.08   /  17.18  &  0.13    /  8.25  &  0.06  /  0.29  &  0.04 /    38.32 &  0.10 /    27.11 &  0.00 /   11.54 \bigstrut\\
		&   16.92 /    25.03 &  8.40       /  14.76    &   0.25       /  0.22     &   40.59       /  68.89    &   24.98 /    59.80    &   11.54       /  16.83    \bigstrut\\
		\hline
		\multirow{2}[4]{*}{ R1-cap to R2-opt} &  0.10   /  17.14 &  0.17 /    10.15 &  0.28 /  0.55  &  0.06     / 44.48 &  0.13      /  33.81 &  0.00 / 7.23  \bigstrut\\
		&   17.97      /  31.73     &  11.35       /  21.07    &   0.61 /    0.56     &   43.22 /    85.73    &   36.51 /    76.07    &   7.23 /    11.44    \bigstrut\\
		\hline
		\multicolumn{6}{c}{\textbf{Reference to Latent}} \bigstrut\\
		\hline
		\multirow{2}[4]{*}{ R1-opt to L-wall} &  19.74  /  49.12 &  30.94   /  50.13 & 33.27 /   27.56 &   20.63    /   53.72 &   19.68     /  50.00 &  4.41 / 50.33 \bigstrut\\
		&   46.87      /  98.85    &  47.06   /  99.47    &   29.34      /  57.98    &   50.78     /  99.15    &  48.76  /  99.58    &   47.53      / 98.32    \bigstrut\\
		\hline
		\multirow{2}[4]{*}{ R1-opt to L-ipad} &  18.22  /  50.00 &  28.67  / 48.60 &  29.15 /  10.23 &  16.60    /   51.96 &  17.92     /   50.00 &  4.17 /   49.13 \bigstrut\\
		&  49.70  /  98.77    &  44.71  /   99.00    &   20.57     /   30.64    &  46.33  /   98.93    &   52.90      /   99.76    &   46.79      /   96.71    \bigstrut\\
		\hline
		\multirow{2}[4]{*}{ R1-opt to L-alum} &  20.64 /  52.32 &  34.38  /   47.15 &  35.49 /   29.43 &  21.36   /   54.22 &  21.08     /   50.00 &  5.04 / 49.89 \bigstrut\\
		&  40.48   /  100.00   &   45.86    /   97.84    &   27.11      /   73.00    &   56.39      /   99.63    &   49.27     /   99.26    &   45.44     /   97.71    \bigstrut\\
		\hline
		\multirow{2}[4]{*}{ R1-cap to L-wall} &  17.99   /  50.65 &  26.48  /  51.17 &  27.31 /  31.33 &  16.45    /  55.26 &  16.55    /  50.00 &  4.06 /  50.89 \bigstrut\\
		&  39.81   /  99.24    &   55.92      /   99.35    &   28.56     /  100.00   &  50.72  /  100.00   &   53.61      /  99.45    &    48.31     /  97.97    \bigstrut\\
		\hline
		\multirow{2}[4]{*}{ R1-cap to L-ipad} &  17.27   /  50.33 &  26.03  /  48.56 &  28.47 /  11.21 &  15.87    /  54.72 &  16.18     /  50.00 &  3.97 /   49.64 \bigstrut\\
		&  40.55   /  99.23    &   46.33     /   99.17    &  22.12  /   55.23    &   50.88      /   99.26    &   46.25      /  99.33    &  45.46  /  97.00    \bigstrut\\
		\hline
		\multirow{2}[4]{*}{ R1-cap to L-alum} &  20.05   /  50.99 &  30.00 /   51.16 &  32.23 /   37.33 &  18.88    /   54.62 &  19.41     /   50.00 &  4.29 /  50.38 \bigstrut\\
		&  40.04   /  100.00   &   54.99     /  97.84    &   33.37     /  90.36    &   59.02     /   98.83    &   51.33    /   99.00    &   48.40     /  98.17    \bigstrut\\
		\hline
		\multicolumn{6}{c}{\textbf{Latent to latent}  } \bigstrut\\
		\hline
		\multirow{2}[4]{*}{ L-wall to L-ipad} &  27.57 /    49.36 &  38.65  /  49.03 &  43.74 /  7.15  &  29.97    /  99.97 &  27.68     /  49.73 &  8.67/  48.14 \bigstrut\\
		&   42.91       /  98.27    &   43.45      / 98.71    &   24.88      /  0.00     &   97.43     /  98.64    &   52.01     /  99.27    &   45.43      / 95.89    \bigstrut\\
		\hline
		\multirow{2}[4]{*}{ L-wall to L-alum} &  27.91   /  56.32 &  39.06  /  52.34 &  44.86 / 16.78 &  31.63    /  99.95 &  28.99     /  50.00 &  9.22/  47.30 \bigstrut\\
		&  55.37   /  100.00   &   48.64      /  98.62    &   27.19      / 50.00    &  96.09  / 99.23    &   49.16      /  98.76    &   44.54      /  94.95    \bigstrut\\
		\hline
		\multirow{2}[4]{*}{ L-ipad to L-alum} &  29.49   /  18.84 &  40.95  / 48.61 &  45.81/ 17.42 &  32.09   / 100.00&  29.92     / 49.13 &  10.27  /  47.09 \bigstrut\\
		&  31.05   /  100.00   &   55.83      / 98.94    &   29.01     /  50.00    &  99.33  / 99.45    &   53.31     / 98.37    &   50.00    / 88.22    \bigstrut\\
		\hline
		\multicolumn{6}{c}{\textbf{Smartphone fingerphoto} } \bigstrut\\
		\hline
		\multirow{2}[4]{*}{ R1-opt to Smt   } &  23.48   /  50.00 &  33.25  /  50.74 &  38.07 / 46.08 &  26.61   / 54.63 &  23.89   /  49.22 &  6.42 /  49.96 \bigstrut\\
		&  46.34   /  98.58    &   45.20      / 98.53    &  42.93  / 97.27    &   49.84     /  99.21    &   50.01    / 99.62    &   45.76     /  98.96    \bigstrut\\
		\hline
		\multirow{2}[4]{*}{ R1-cap to Smt   } &  23.17 /    50.00 &  33.02  / 48.53 &  36.94 /  17.86 &  25.43   /  54.31 &  22.58    / 50.00 &  6.05 / 50.00 \bigstrut\\
		&   42.19 /    99.22    &   49.08     /  100.00   &   30.92     /  98.33    &   50.79     /  98.57    &   51.84    / 98.92    &   47.00     /  98.51    \bigstrut\\
		\hline
		\multirow{2}[4]{*}{ Smt to L-wall   } &  25.09 /    50.00 &  35.46  / 48.19 &  39.85 / 55.85 &  28.00    / 100.00&  25.84    / 49.78 &  7.24 /  49.44 \bigstrut\\
		&  54.81   /  98.73    &   43.84      / 98.16    &   48.71      / 80.00    &   98.73     /  99.42    &   48.74     /  99.55    &   50.00     / 99.00    \bigstrut\\
		\hline
		\multirow{2}[4]{*}{ Smt to L-ipad   } &  25.89   /  49.74 &  36.04 /  52.46 &  40.37 / 52.25 &  29.44   / 99.78 &  26.38    / 49.43 &  7.55 / 49.84 \bigstrut\\
		&  47.35   /  98.87    &   55.70      / 98.31    &   44.33     /  97.26    &   96.24    /  98.55    &   45.19    / 99.72    &   44.44    /  98.51    \bigstrut\\
		\hline
		\multirow{2}[4]{*}{ Smt to L-alum   } &  26.87   /  50.00 &  36.95  / 55.25 &  40.86 / 55.52 &  30.08   /  99.98 &  26.75    / 50.00 &  7.68 / 49.84 \bigstrut\\
		&  52.74 /    100.00   &   49.89     / 100.00   &   57.05      / 100.00   &   97.07      / 99.12    &   53.03     /  99.16    &   46.38   /  99.19    \bigstrut\\
		\hline
	\end{tabular}%
}
\label{tab:5-errors-results}
\end{table*}%        

\subsection{Protocol III: Verification results - Fingerphoto Comparisons}

We further consider another protocol according to recent trends and benchmark the performance for fingerphoto to latent fingerprint comparison. Specifically, the protocol is aimed at using fingerphotos as a replacement to traditional fingerprint capture from contact-based sensors. We therefore evaluate, fingerphotos as reference and compare it to latent comparisons. From Fig. \ref{fig:5-det-results} and Table \ref{tab:5-errors-results} (in the bottom sector), one can note that the EER (more than 50\%) and FMR100 are high and indicate a challenging nature of smartphone fingerphotos to latent comparison in the LFIW database.

\begin{figure*}[]
\centering
  \centering

 \begin{minipage}[b]{0.75\textwidth} 
     \begin{subfigure}[b]{0.32\textwidth}
         \centering
         \includegraphics[width=\textwidth]{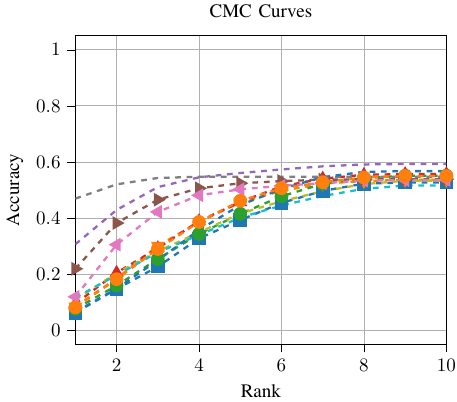}
         \caption{NBIS}
     \end{subfigure}
     \hfill
     \begin{subfigure}[b]{0.32\textwidth}
         \centering
         \includegraphics[width=\textwidth]{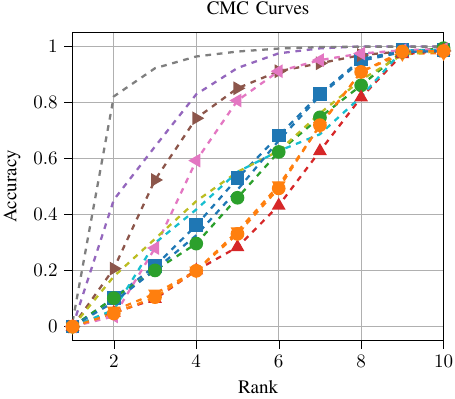}
         \caption{MCC}
     \end{subfigure}
     \hfill
     \begin{subfigure}[b]{0.32\textwidth}
         \centering
         \includegraphics[width=\textwidth]{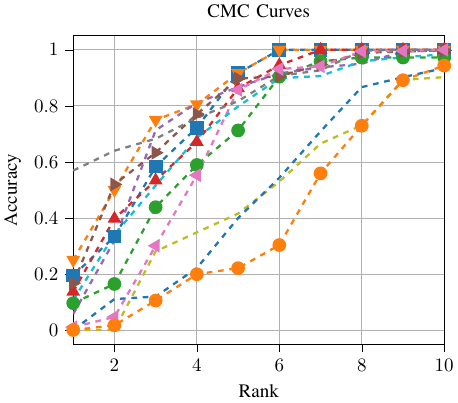}
         \caption{VeriFinger}
     \end{subfigure}
     \hfill
     \begin{subfigure}[b]{0.32\textwidth}
         \centering
        \includegraphics[width=\textwidth]{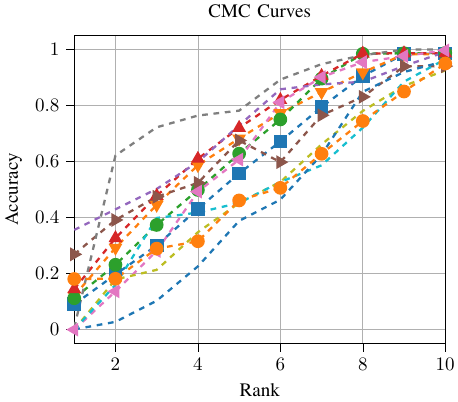}
         \caption{MN-MCC}
     \end{subfigure}
     \hfill
     \begin{subfigure}[b]{0.32\textwidth}
         \centering
         \includegraphics[width=\textwidth]{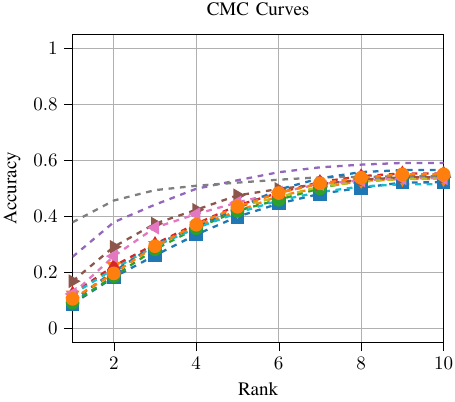}
         \caption{MN-NBIS}
     \end{subfigure}
     \hfill
     \begin{subfigure}[b]{0.32\textwidth}
         \centering
         \includegraphics[width=\textwidth]{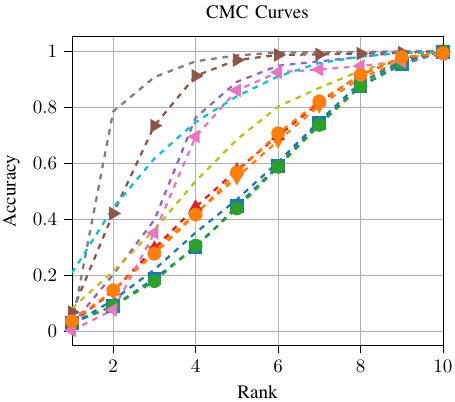}
         \caption{MSU-LAFIS}
     \end{subfigure}
\end{minipage}
\begin{minipage}[b]{0.23\textwidth}
\centering
    \includegraphics[width=.8\textwidth]{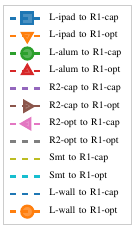}
    \caption*{Legend}
\end{minipage}
     
        \caption{Cumulative match curves of the rank identification rates for NBIS, MCC, VeriFinger, MinutiaeNet-MCC, MinutiaeNet-NBIS, and MSU-LAFIS identification experiments for the LFIW database}
        \label{fig:5-rank-results}
\end{figure*}

\subsection{Identification results}

In addition to demonstrating verification results, we also illustrate the performance of identification results using Cumulative Match Curves (CMC) of the Rank Identification (RI) rates (rank-10) for NBIS, MCC, VeriFinger, MinutiaeNet-MCC, MinutiaeNet-NBIS, and MSU-LAFIS identification experiments for the LFIW database as shown in Fig. \ref{fig:5-rank-results}. From Fig. \ref{fig:5-rank-results} we can observe that the overall RI rates for reference comparisons are higher than latent identification. The RI rates for MCC, VeriFinger, MinutiaeNet-MCC, and MSU-LAFIS can reach 100\% at rank-6 to rank-10. While the RI rates for NBIS and MinutiaeNet-NBIS are lower than 60\% in all 10 ranks. Similar to the verification results, 'R2-opt to R1-opt' comparisons (gray dashed lines in Fig. \ref{fig:5-rank-results}) remains the best performance but none of the algorithms can get RI rates to 1 before rank-4.

In summary, none of the selected algorithms can provide high accuracy latent fingerprints minutiae extraction and comparison by using fingerprints from the LFIW database for both verification and identification scenarios. From the experimental results above we can conclude that latent fingerprint recognition by existing techniques is still very challenging and complex, especially for latent fingerprints captured in the wild conditions. The results indicate again the need for the development of robust latent fingerprint minutiae extraction and recognition algorithms for wild latent fingerprints. %The ROC curve, FMR and FNMR curves, distributions of the comparison scores, and the full performance indicators are illustrated in Section G of the Supplementary Material (Fig. 66-75, and Table XXXIII-XXXVI). 

% \subsection{Score-level fusion}

% We attempt to use comparison score level fusion techniques to verify if the system performance can be improved. Comparison score level fusion is one of the commonly used because comparison scores are easily obtained and include rich information to distinguish between a mated and an impostor trait. Therefore, the fusion of the comparison scores from the selected algorithms seems both feasible and practical for the improvement of system performance. Three different fusion strategies are tested: 1) simple sum rule, 2) weighted sum fusion with score normalization, and 3) SVM classifier based fusion. We use $30\%$ of the comparison scores for computing the normalization parameters and SVM training. Different combinations of comparison algorithms are tested to achieve the best fusion results. For simple sum rule score fusion, the best system performance is $EER=0.3975$ by combining all the selected six algorithms (both MCC and NBIS are used for MinutiaeNet). For weighted sum fusion with score normalization, the lowest EER is $0.2833$ and it comes from the combination of NBIS, MCC, MinutiaeNet-MCC, and MSU. 

\subsection{Directions for future works}

As illustrated from the experimental results discussed in the above sections, the system performance of evaluated latent fingerprints recognition algorithms does not meet the operational requirements. By looking at FMR100, we can see from the overall results in Table \ref{tab:all-errors-results} or the reference/latent comparisons results in Table \ref{tab:5-errors-results} that the result is around 20\% only for reference fingerprints comparison experiments. For latent and smartphone fingerprints comparison experiments, the FMR100 is higher than 90\% and many of them are even close to 100\%. From a practical point of view, this behaviour would cause a noticeable number of false matches and, as a result, a bigger number of false non matches during latent fingerprints recognition at crime scene, digital forensic, or law enforcement scenarios. This would not help to find the real suspects. Therefore, the directions for the future works could be as the following:
\begin{itemize}
    \item Given that the existing fingerprint recognition systems have very low performance on smartphone and latent fingerprints captured from different material, more accurate and robust algorithms are desired in order to overcome the difficulties and challenges of wild latent fingerprints recognition.
    \item As it also has been discussed that the FTER is relatively high for existing fingerprint recognition algorithms, including systems that particularly for latent fingerprints. Reliable and accurate latent fingerprints pre-processing (e.g.segmentation, quality enhancement, minutiae orientation estimation, etc.) and minutiae extraction approaches need to be developed. 
    \item While there exist quality assessment algorithms like NFIQ \cite{ISO-IEC-29794-4} for fingerprints, quality assessment algorithms that can predict the recognition performance for latent fingerprints are still missing. 
    \item As an additional direction, the performance of human examiner latent fingerprints comparison could be investigated in a standardized manner to discover the important factors in recognizing the latent fingerprints captured in wild environment.
\end{itemize}

\section{Conclusion}
\label{sec:conclusion}

Latent fingerprints recognition has always been a complex and challenging task with the availability of no public and large-scale datasets. In this work, we have introduced LFIW, a new database of latent fingerprints in the wild. This database has included six different acquisition scenarios: reference fingerprints from (1) optical and (2) capacitive traditional fingerprint sensors, (3) smartphone fingerphotos, latent fingerprints captured from (4) wall surface, (5) Ipad surface, and (6) aluminium foil surface. The new database consists of 1318 unique fingerprint instances captured in all of the above-mentioned settings. A total of 2636 reference fingerprints from optical and capacitive sensors, 1318 fingerphotos from smartphone, and 9224 latent fingerprints from every 132 subjects are provided in this work. The presented wild latent fingerprints database with large number of unique fingerprints will be publicly available in order to allow researchers to benchmark their algorithms in a free and sustainable manner to develop robust and accurate latent fingerprints recognition algorithms. Additionally, a benchmark of several existing state-of-the-art fingerprint recognition systems is also provided in this paper to eliminate the limitations in the existing latent fingerprints recognition methods, and to provide some directions for future works in this research field.

% a new evaluation pipeline and
\balance
\section*{Acknowledgments}
This paper is supported by the National Natural Science Foundation of China (Grant No. 62106228, 61906170), Zhejiang Natural Science Foundation (Grant No. LQ22F020003), Ningbo Natural Science Foundation (Grant No. 2021J175), and the Ningbo Yongjiang Talent Introduction Programme 2021.

% {\appendix[Proof of the Zonklar Equations]
% Use $\backslash${\tt{appendix}} if you have a single appendix:
% Do not use $\backslash${\tt{section}} anymore after $\backslash${\tt{appendix}}, only $\backslash${\tt{section*}}.
% If you have multiple appendixes use $\backslash${\tt{appendices}} then use $\backslash${\tt{section}} to start each appendix.
% You must declare a $\backslash${\tt{section}} before using any $\backslash${\tt{subsection}} or using $\backslash${\tt{label}} ($\backslash${\tt{appendices}} by itself
%  starts a section numbered zero.)}

% {
% \appendix

% Decomposed experimental results for NBIS, MCC, VeriFinger, MinutiaeNet, and MSU-LAFIS are illustrated here in Table \ref{tab:nbis-errors-appendix}, \ref{tab:mcc-errors-appendix}, \ref{tab:verifinger-errors-appendix}, \ref{tab:minutiaenet-mcc-errors-appendix}, \ref{tab:minutiaenet-nbis-errors-appendix}, and \ref{tab:msu-errors-appendix}.
% }

\bibliography{lfiw-tifs-2023-230403.bib}

\begin{thebibliography}{10}
\providecommand{\url}[1]{#1}
\csname url@samestyle\endcsname
\providecommand{\newblock}{\relax}
\providecommand{\bibinfo}[2]{#2}
\providecommand{\BIBentrySTDinterwordspacing}{\spaceskip=0pt\relax}
\providecommand{\BIBentryALTinterwordstretchfactor}{4}
\providecommand{\BIBentryALTinterwordspacing}{\spaceskip=\fontdimen2\font plus
\BIBentryALTinterwordstretchfactor\fontdimen3\font minus
  \fontdimen4\font\relax}
\providecommand{\BIBforeignlanguage}[2]{{%
\expandafter\ifx\csname l@#1\endcsname\relax
\typeout{** WARNING: IEEEtran.bst: No hyphenation pattern has been}%
\typeout{** loaded for the language `#1'. Using the pattern for}%
\typeout{** the default language instead.}%
\else
\language=\csname l@#1\endcsname
\fi
#2}}
\providecommand{\BIBdecl}{\relax}
\BIBdecl

\bibitem{cao2018automated}
K.~Cao and A.~K. Jain, ``Automated latent fingerprint recognition,'' \emph{IEEE
  transactions on pattern analysis and machine intelligence}, vol.~41, no.~4,
  pp. 788--800, 2018.

\bibitem{maltoni2022latent}
D.~Maltoni, D.~Maio, A.~K. Jain, and J.~Feng, ``Latent fingerprint
  recognition,'' \emph{Handbook of Fingerprint Recognition}, pp. 339--383,
  2022.

\bibitem{fbi2020}
``Fbi biometric services section, june 2021 next generation identification
  (ngi) system fact sheet, 2021,''
  https://www.fbi.gov/file-repository/ngi-monthly-fact-sheet., accessed:
  2022-11-16.

\bibitem{fpvte}
``National institute of standards \& technology,''
  https://www.nist.gov/programs-projects/fingerprint-vendor-technology-evaluation-fpvte,
  accessed: 2022-11-09.

\bibitem{watson2012fingerprint}
C.~Watson, G.~Fiumara, E.~Tabassi, S.~Cheng, P.~Flanagan, and W.~Salamon,
  ``Fingerprint vendor technology evaluation: Evaluation of fingerprint
  matching algorithms,'' \emph{NISTIR}, vol. 8034, p. 233, 2012.

\bibitem{indovina2012evaluation}
M.~D. Indovina, V.~Dvornychenko, R.~Hicklin, and G.~Kiebuzinski,
  \emph{Evaluation of latent fingerprint technologies: Extended feature sets
  [evaluation\# 2]}, 2012.

\bibitem{singla2020automated}
N.~Singla, M.~Kaur, and S.~Sofat, ``Automated latent fingerprint identification
  system: A review,'' \emph{Forensic science international}, vol. 309, p.
  110187, 2020.

\bibitem{garris2000nist}
M.~D. Garris and M.~D. Garris, \emph{NIST special database 27: Fingerprint
  minutiae from latent and matching tenprint images}.\hskip 1em plus 0.5em
  minus 0.4em\relax US Department of Commerce, National Institute of Standards
  and Technology, 2000.

\bibitem{sankaran2011matching}
A.~Sankaran, T.~I. Dhamecha, M.~Vatsa, and R.~Singh, ``On matching latent to
  latent fingerprints,'' in \emph{2011 international joint conference on
  biometrics (IJCB)}.\hskip 1em plus 0.5em minus 0.4em\relax IEEE, 2011, pp.
  1--6.

\bibitem{feng2012robust}
J.~Feng, Y.~Shi, and J.~Zhou, ``Robust and efficient algorithms for separating
  latent overlapped fingerprints,'' \emph{IEEE Transactions on Information
  Forensics and Security}, vol.~7, no.~5, pp. 1498--1510, 2012.

\bibitem{ICPSR2009}
``Census of publicly funded forensic crime lab.'' https://www.
  icpsr.umich.edu/web/ICPSR/studies/34340, accessed: 2022-11-09.

\bibitem{wvudatabase}
iPRoBe Lab, ``{WVU Latent Fingerprint Database},'' http://www.csee.wvu.edu/
  ross/i-probe/.

\bibitem{sankaran2015multisensor}
A.~Sankaran, M.~Vatsa, and R.~Singh, ``Multisensor optical and latent
  fingerprint database,'' \emph{IEEE access}, vol.~3, pp. 653--665, 2015.

\bibitem{sankaran2012hierarchical}
A.~Sankaran, M.~Vatsa, and R.~Singh., ``Hierarchical fusion for matching
  simultaneous latent fingerprint,'' in \emph{2012 IEEE Fifth International
  Conference on Biometrics: Theory, Applications and Systems (BTAS)}.\hskip 1em
  plus 0.5em minus 0.4em\relax IEEE, 2012, pp. 377--382.

\bibitem{paulino2012latent}
A.~A. Paulino, J.~Feng, and A.~K. Jain, ``Latent fingerprint matching using
  descriptor-based hough transform,'' \emph{IEEE Transactions on Information
  Forensics and Security}, vol.~8, no.~1, pp. 31--45, 2012.

\bibitem{sankaran2015latent}
A.~Sankaran, A.~Agarwal, R.~Keshari, S.~Ghosh, A.~Sharma, M.~Vatsa, and
  R.~Singh, ``Latent fingerprint from multiple surfaces: Database and quality
  analysis,'' in \emph{7th International Conference on Biometrics Theory,
  Applications and Systems (BTAS)}.\hskip 1em plus 0.5em minus 0.4em\relax
  IEEE, 2015, pp. 1--6.

\bibitem{zhang2012latent}
J.~Zhang, R.~Lai, and C.-C.~J. Kuo, ``Latent fingerprint detection and
  segmentation with a directional total variation model,'' in \emph{19th
  International Conf. on Image Processing}.\hskip 1em plus 0.5em minus
  0.4em\relax IEEE, 2012, pp. 1145--1148.

\bibitem{cao2014segmentation}
K.~Cao, E.~Liu, and A.~K. Jain, ``Segmentation and enhancement of latent
  fingerprints: A coarse to fine ridgestructure dictionary,'' \emph{IEEE
  transactions on pattern analysis and machine intelligence}, vol.~36, no.~9,
  pp. 1847--1859, 2014.

\bibitem{stojanovic2016fingerprint}
B.~Stojanovi{\'c}, O.~Marques, A.~Ne{\v{s}}kovi{\'c}, and S.~Puzovi{\'c},
  ``Fingerprint roi segmentation based on deep learning,'' in \emph{2016 24th
  Telecommunications Forum (TELFOR)}.\hskip 1em plus 0.5em minus 0.4em\relax
  IEEE, 2016, pp. 1--4.

\bibitem{sankaran2017adaptive}
A.~Sankaran, A.~Jain, T.~Vashisth, M.~Vatsa, and R.~Singh, ``Adaptive latent
  fingerprint segmentation using feature selection and random decision forest
  classification,'' \emph{Information Fusion}, vol.~34, pp. 1--15, 2017.

\bibitem{nguyen2018automatic}
D.-L. Nguyen, K.~Cao, and A.~K. Jain, ``Automatic latent fingerprint
  segmentation,'' in \emph{2018 IEEE 9th International Conference on Biometrics
  Theory, Applications and Systems (BTAS)}.\hskip 1em plus 0.5em minus
  0.4em\relax IEEE, 2018, pp. 1--9.

\bibitem{chen2011separating}
F.~Chen, J.~Feng, A.~K. Jain, J.~Zhou, and J.~Zhang, ``Separating overlapped
  fingerprints,'' \emph{IEEE Transactions on Information Forensics and
  Security}, vol.~6, no.~2, pp. 346--359, 2011.

\bibitem{zhao2012model}
Q.~Zhao and A.~K. Jain, ``Model based separation of overlapping latent
  fingerprints,'' \emph{IEEE Transactions on Information Forensics and
  Security}, vol.~7, no.~3, pp. 904--918, 2012.

\bibitem{jeyanthi2016efficient}
S.~Jeyanthi, N.~Uma~Maheswari, and R.~Venkatesh, ``An efficient automatic
  overlapped fingerprint identification and recognition using anfis
  classifier,'' \emph{International Journal of Fuzzy Systems}, vol.~18, no.~3,
  pp. 478--491, 2016.

\bibitem{stojanovic2017novel}
B.~Stojanovi{\'c}, A.~Ne{\v{s}}kovi{\'c}, and O.~Marques, ``A novel neural
  network based approach to latent overlapped fingerprints separation,''
  \emph{Multimedia Tools and Applications}, vol.~76, no.~10, pp.
  12\,775--12\,799, 2017.

\bibitem{su2010latent}
C.~Su and S.~Srihari, ``Latent fingerprint core point prediction based on
  gaussian processes,'' in \emph{2010 20th International Conference on Pattern
  Recognition}.\hskip 1em plus 0.5em minus 0.4em\relax IEEE, 2010, pp.
  1634--1637.

\bibitem{sankaran2014latent}
A.~Sankaran, P.~Pandey, M.~Vatsa, and R.~Singh, ``On latent fingerprint
  minutiae extraction using stacked denoising sparse autoencoders,'' in
  \emph{IEEE International Joint Conference on Biometrics}.\hskip 1em plus
  0.5em minus 0.4em\relax IEEE, 2014, pp. 1--7.

\bibitem{tang2017latent}
Y.~Tang, F.~Gao, and J.~Feng, ``Latent fingerprint minutia extraction using
  fully convolutional network,'' in \emph{2017 IEEE International Joint
  Conference on Biometrics (IJCB)}.\hskip 1em plus 0.5em minus 0.4em\relax
  IEEE, 2017, pp. 117--123.

\bibitem{jain2010latent}
A.~K. Jain and J.~Feng, ``Latent fingerprint matching,'' \emph{IEEE
  Transactions on pattern analysis and machine intelligence}, vol.~33, no.~1,
  pp. 88--100, 2010.

\bibitem{nguyen2019end}
D.-L. Nguyen and A.~K. Jain, ``End-to-end pore extraction and matching in
  latent fingerprints: Going beyond minutiae,'' \emph{arXiv preprint
  arXiv:1905.11472}, 2019.

\bibitem{maio2002fvc2002}
D.~Maio, D.~Maltoni, R.~Cappelli, J.~L. Wayman, and A.~K. Jain, ``{FVC2002:
  Second fingerprint verification competition},'' in \emph{International Conf.
  on Pattern Recognition}, vol.~3.\hskip 1em plus 0.5em minus 0.4em\relax IEEE,
  2002, pp. 811--814.

\bibitem{cappelli2007fingerprint}
R.~Cappelli, M.~Ferrara, A.~Franco, and D.~Maltoni, ``Fingerprint verification
  competition 2006,'' \emph{Biometric Technology Today}, vol.~15, no. 7-8, pp.
  7--9, 2007.

\bibitem{nbis}
``{NIST Biometric Image Software},''
  https://www.nist.gov/services-resources/software/nist-biometric-image-software-nbis,
  accessed: 2022-12-25.

\bibitem{mcc}
``{Minutia Cylinder-Code},'' http://biolab.csr.unibo.it/Home.asp, accessed:
  2022-12-25.

\bibitem{cappelli2010minutia}
R.~Cappelli, M.~Ferrara, and D.~Maltoni, ``Minutia cylinder-code: A new
  representation and matching technique for fingerprint recognition,''
  \emph{IEEE transactions on pattern analysis and machine intelligence},
  vol.~32, no.~12, pp. 2128--2141, 2010.

\bibitem{iso2011iec}
{ISO/IEC JTC1 SC37 Biometrics}, \emph{{IEC 19794-2: 2011 Information
  technology--Biometric data interchange formats--Part 2: Finger minutiae
  data}}, 2011.

\bibitem{dorizzi2009fingerprint}
B.~Dorizzi, R.~Cappelli, M.~Ferrara, D.~Maio, D.~Maltoni, N.~Houmani,
  S.~Garcia-Salicetti, and A.~Mayoue, ``Fingerprint and on-line signature
  verification competitions at icb 2009,'' in \emph{International Conference on
  Biometrics}.\hskip 1em plus 0.5em minus 0.4em\relax Springer, 2009, pp.
  725--732.

\bibitem{biolab}
``{Fvc-Ongoing Web Site. [Online].}'' https://biolab.csr.unibo.it/fvcongoing/
  \\ UI/Form/BenchmarkAreas/BenchmarkAreaDMAD.aspx, accessed: 2022-12-19.

\bibitem{verifinger}
``Verifinger sdk,'' https://www.neurotechnology.com/verifinger.html, accessed:
  2022-12-25.

\bibitem{Nguyen_MinutiaeNet}
D.-L. Nguyen, K.~Cao, and A.~K. Jain, ``Robust minutiae extractor: Integrating
  deep networks and fingerprint domain knowledge,'' in \emph{The 11th
  International Conference on Biometrics, 2018}, 2018.

\bibitem{cao2019end}
K.~Cao, D.-L. Nguyen, C.~Tymoszek, and A.~K. Jain, ``End-to-end latent
  fingerprint search,'' \emph{IEEE Transactions on Information Forensics and
  Security}, vol.~15, pp. 880--894, 2019.

\bibitem{ISO-IEC-29794-4}
{ISO/IEC JTC1 SC37 Biometrics}, \emph{{ISO/IEC} 29794-4. Information technology
  — Biometric sample quality — Part 4: Finger image data}, International
  Organization for Standardization, 2017.

\end{thebibliography}

% \newpage

 
\vspace{11pt}

% \end{IEEEbiography}

% \end{IEEEbiography}



% \newpage

% \includepdf[pages=-]{SUPPLEMENTARY MATERIAL.pdf}

\vfill

\end{document}